%% file: main.tex
\documentclass[10pt,twocolumn,letterpaper]{article}

\usepackage{cvpr}              % To produce the CAMERA-READY version

\usepackage{multicol}
\usepackage{multirow}
\usepackage{subcaption}
\usepackage{balance} 
\usepackage{xcolor}
\usepackage{pifont} % 导言区添加
\newcommand{\cmark}{\ding{51}} % 定义勾号符号
\usepackage{booktabs}
\usepackage{multirow}
\usepackage{graphicx}
\usepackage[table]{xcolor} % 支持灰底
\usepackage{dblfloatfix}
\usepackage{arydshln}
\usepackage[linesnumbered,ruled,vlined]{algorithm2e}
\usepackage{float}
\definecolor{darkgreen}{RGB}{0,100,0} % 深绿色

\usepackage{bibunits}
\defaultbibliographystyle{ieeenat_fullname}

\input{preamble}
\definecolor{cvprblue}{rgb}{0.21,0.49,0.74}
\usepackage[pagebackref,breaklinks,colorlinks,allcolors=cvprblue]{hyperref}

\def\paperID{} % *** Enter the Paper ID here
\def\confName{CVPR}
\def\confYear{2026}

\title{FB-CLIP: Fine-Grained Zero-Shot Anomaly Detection with Foreground-Background Disentanglement}

\author{
    Ming Hu$^{1,2}$ \qquad 
    Yongsheng Huo$^{1,2}$ \qquad 
    Mingyu Dou$^{1,2}$ \qquad 
    Jianfu Yin$^{1,2}$ \qquad 
    Peng Zhao$^{1,2}$ \qquad \\
    Yao Wang$^{3}$ \qquad
     Cong Hu$^{4}$\thanks{Corresponding author.} \qquad 
    Bingliang Hu$^{1}$\footnotemark[1] \qquad 
    Quan Wang$^1$\footnotemark[1] \\
    $^1$ Xi'an Institute of Optics and Precision Mechanics, Chinese Academy of Sciences \\
    $^2$ University of Chinese Academy of Sciences \quad 
    $^3$ Xi'an Jiaotong University\\
   $^4$  Zhongnan Hospital of Wuhan University
}

\begin{document}
\maketitle

\input{sec/0_abstract}    
\input{sec/1_intro}
\input{sec/2_related}
\input{sec/3_method}
\input{sec/4_exe}
\input{sec/5_condlusion}

\clearpage

\section*{Acknowledgment}
This research was supported by:
The Shaanxi Province Technological Innovation Guidance Special Project: Regional Science and Technology Innovation Center, Strategic Scientific and Technological Strength Category  [No.2024QY-SZX-26].
The Hubei Provincial Natural Science Foundation (Grant No. JCZRLH202600590).
The research was supported by the Key Laboratory of Spectral Imaging Technology, Xi'an Institute of Optics and Precision Mechanics of the Chinese Academy of Sciences; the Key Laboratory of Biomedical Spectroscopy of Xi'an.
{
    \small
    \bibliographystyle{ieeenat_fullname}
    \bibliography{main}
}

\input{sec/X_suppl}

\end{document}

%% file: sec/0_abstract.tex
\begin{abstract}

Fine-grained anomaly detection is crucial in industrial and medical applications, but labeled anomalies are often scarce, making zero-shot detection challenging. While vision-language models like CLIP offer promising solutions, they struggle with foreground-background feature entanglement and coarse textual semantics.  
We propose FB-CLIP, a framework that enhances anomaly localization via multi-strategy textual representations and foreground-background separation. In the textual modality, it combines End-of-Text features, global-pooled representations, and attention-weighted token features for richer semantic cues. In the visual modality, multi-view soft separation along identity, semantic, and spatial dimensions, together with background suppression, reduces interference and improves discriminability. Semantic Consistency Regularization (SCR) aligns image features with normal and abnormal textual prototypes, suppressing uncertain matches and enlarging semantic gaps.  
Experiments show that FB-CLIP effectively distinguishes anomalies from complex backgrounds, achieving accurate fine-grained anomaly detection and localization under zero-shot settings.
Code  : \href{https://github.com/Xi-Mu-Yu/FB-CLIP}{https://github.com/Xi-Mu-Yu/FB-CLIP}.

\end{abstract}

%% file: sec/1_intro.tex
\section{Introduction}
\label{sec:intro}

\begin{figure}[htbp]
\centering
\includegraphics[width=0.5\textwidth]{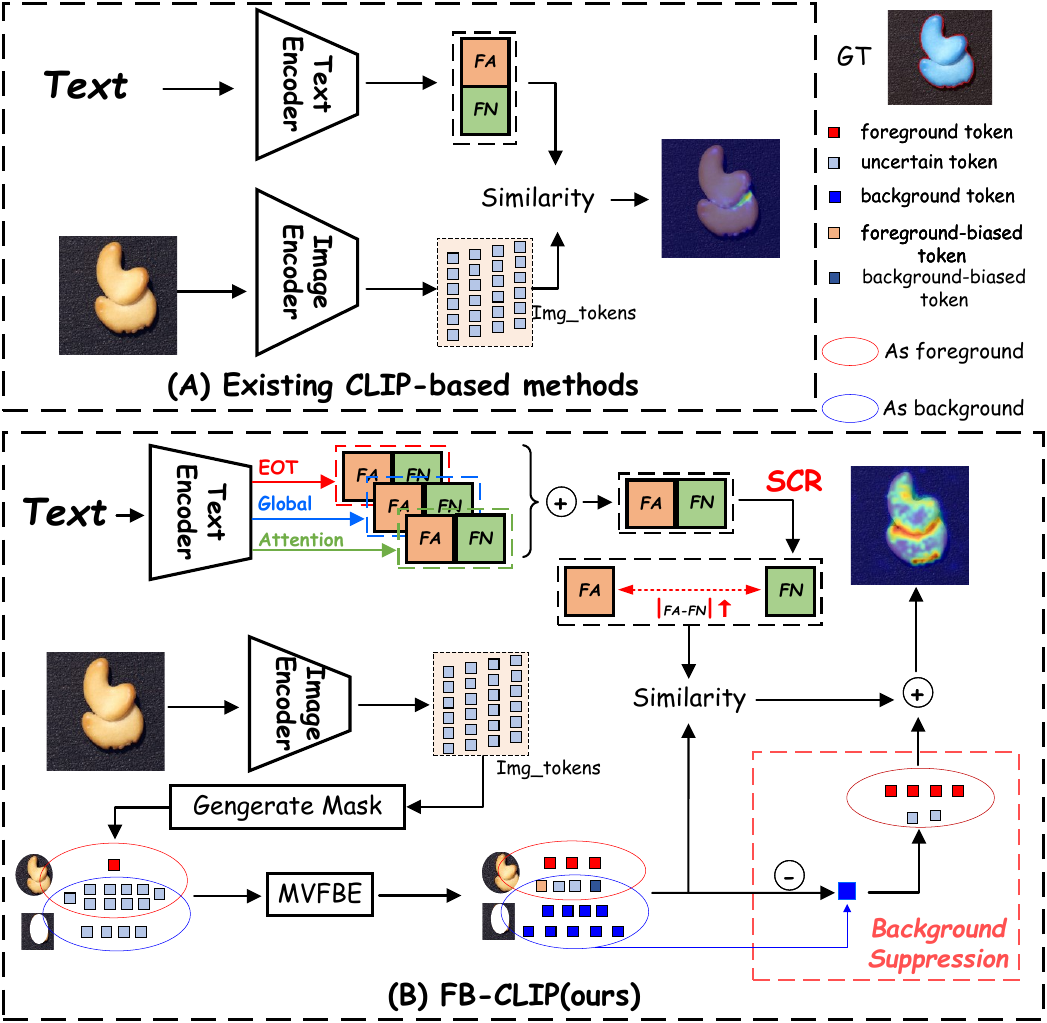}
\caption{Framework comparison. (a) Existing methods use a single text feature for uniform alignment, leading to coarse semantics. (b) FB-CLIP employs multi-text features and Semantic Consistency Regularization (SCR) to enhance alignment, while foreground–background separation and background suppression enable fine-grained, noise-robust anomaly representation.}
\label{fig:pic1}
\end{figure}

Anomaly detection (AD) aims to identify data samples or regions that deviate significantly from normal patterns, and plays a critical role in applications ranging from industrial inspection~\citep{bergmann2019mvtec, xie2023pushing, roth2022towards, huang2022registration, mou2022rgi, chen2022deep, bergmann2020uninformed, pang2021explainable, reiss2023mean, you2022unified, liznerski2020explainable, ding2022catching, 9940966, cao2023anomaly, zhou2023anomalyclip} to medical imaging~\citep{pang2021explainable, qin2022medical, liu2023clip, ding2022catching, tian2021constrained, tian2023self, fernando2021deep}. Despite its importance, labeled anomalous data are often scarce or unavailable~\cite{hu2025beta}, motivating research into unsupervised and zero-shot anomaly detection (ZSAD)~\cite{zhou2023anomalyclip, jeong2023winclip}. 
Recent advances in vision–language pretraining models (VLMs)~\citep{radford2021learning, kirillov2023segment}, particularly CLIP~\citep{radford2021learning}, have enabled semantic alignment between visual and textual modalities, opening new opportunities for ZSAD.

\begin{figure}[htpb]
\centering
\includegraphics[width=0.4\textwidth]{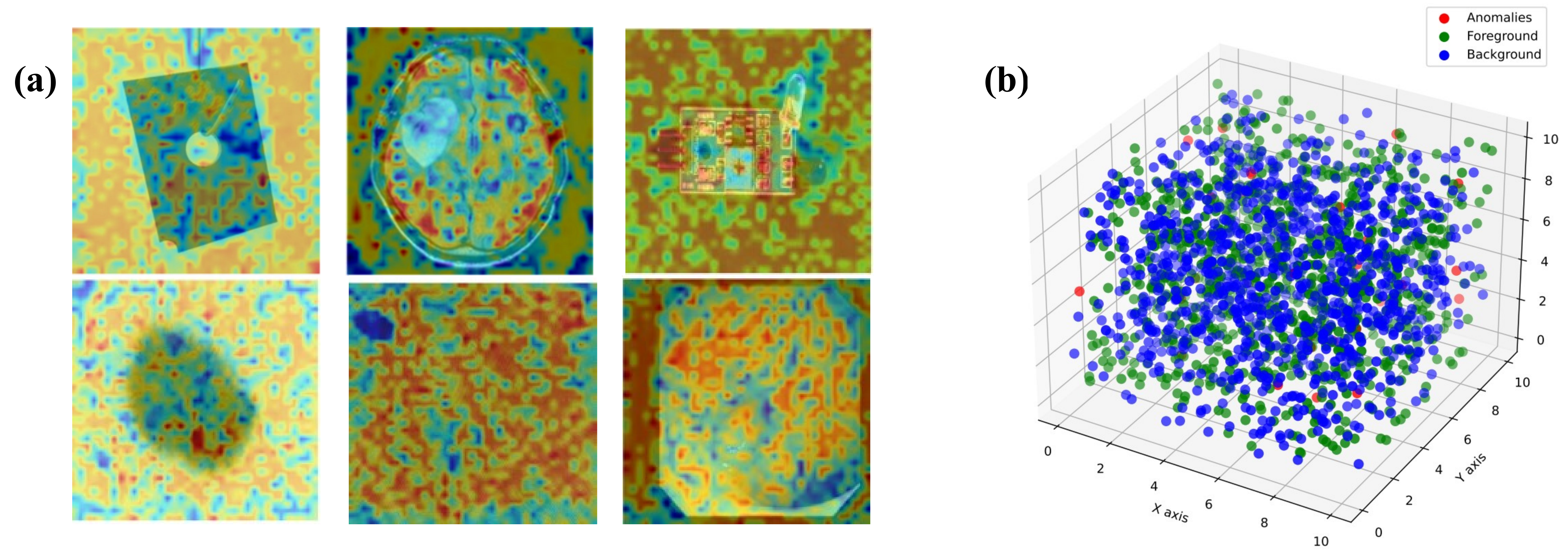}
\caption{(a) Visualization of the responses from the original CLIP model, corresponding to Fig.~1(c) in AnomalyCLIP~\cite{zhou2023anomalyclip}. (b) Entanglement of foreground, background, and anomalous regions. Separating foreground and background clarifies the distinction between anomalous and normal points, improving accuracy.}
\label{fig:pic_map}
\end{figure}

Although CLIP has opened a promising direction for ZSAD, directly applying it to fine-grained anomaly detection remains challenging. By examining the visualization of the original CLIP responses presented in AnomalyCLIP (see Fig.~\ref{fig:pic_map} (a)), we observe CLIP often produces strong responses in both foreground and background regions simultaneously. This indicates that the model struggles to distinguish anomaly-related foreground semantics from irrelevant background context, resulting in pronounced foreground–background feature entanglement in visual representations~\cite{li2024densevlm, yao2026disa, han2025alignclip}, as illustrated in Fig.~\ref{fig:pic_map} (b). In such cases, abundant background responses may obscure subtle anomaly signals, making it difficult for the model to accurately perceive and localize fine-grained anomalies.

As illustrated in Fig.~\ref{fig:pic1}(a), existing CLIP-based anomaly detection methods attempt to alleviate this issue mainly from two perspectives: text modeling and visual feature enhancement. On the textual side~\citep{jeong2023winclip, zhou2023anomalyclip, cao2024adaclip, fang2025af, zhu2025fine, qu2025bayesian}, many approaches introduce learnable prompts or manually designed descriptions to project textual semantics into the shared embedding space and align them with image features. Representative works such as AnomalyCLIP attempt to learn object-agnostic prompts that encode generic normal and abnormal semantics. While these strategies improve cross-domain generalization to some extent, the textual representations themselves often remain relatively coarse and limited in semantic diversity. As a result, the textual features may provide insufficient guidance for distinguishing foreground anomaly cues from complex background contexts. In fine-grained anomaly scenarios, where abnormal patterns are often subtle and sparsely distributed, such limited semantic expressiveness can lead to ambiguous cross-modal alignment, making it difficult for the model to clearly separate anomaly-related regions from irrelevant background responses.
On the visual side, existing methods typically enhance image tokens to improve patch-level feature representation and alignment~\citep{jeong2023winclip,  cao2024adaclip, fang2025af,  gao2025adaptclip}. However, these approaches usually treat tokens in a homogeneous manner, implicitly assuming that all tokens contribute equally to anomaly representation. This assumption overlooks the large uncertainty and diversity of semantic information across different image regions. In fine-grained anomaly scenarios, some tokens may contain critical anomaly cues, while others mainly capture background context or noise. Without explicitly modeling such differences, the enhanced representations may still suffer from foreground–background ambiguity, which dilutes anomaly responses and reduces localization accuracy. 

To address the foreground–background ambiguity in CLIP-based anomaly detection, as illustrated in Fig.~\ref{fig:pic1}(b), we propose FB-CLIP, a framework designed to improve anomaly localization through foreground–background separation guided by richer semantic representations.
In the textual modality, FB-CLIP constructs more expressive textual features by integrating End-of-Text (EOT) features, global-pooled representations, and attention-weighted token features. Compared with conventional prompt-based text representations, this multi-strategy feature fusion provides more comprehensive semantic cues about normality and abnormality. The richer textual semantics serve as stronger guidance for cross-modal alignment, enabling the model to better differentiate anomaly-related foreground signals from background context during feature matching.
To further stabilize cross-modal alignment and improve fine-grained anomaly localization under zero-shot settings, we introduce Semantic Consistency Regularization (SCR). SCR aligns image features with both normal and abnormal textual prototypes, and by minimizing prediction entropy and imposing margin constraints between semantic prototypes, it suppresses uncertain matches and enlarges semantic gaps. This enhances semantic discriminability within the single-text-space alignment, further improving anomaly localization performance.
In the \textbf{visual modality}, FB-CLIP performs multi-view foreground–background soft separation along \textit{identity}, \textit{semantic}, and \textit{spatial} dimensions to refine token representations. Continuous gating integrates dual-branch attention outputs, preventing information loss caused by hard filtering. Moreover, the Background Suppression (BS) module leverages multi-layer, multi-scale background prototypes to further reduce residual interference, mitigating the impact of complex scenes and cross-domain distribution shifts. These designs enable FB-CLIP to produce cleaner and more discriminative anomaly representations, facilitating accurate detection and localization.
Our contributions are as follows:

\textbf{Multi-strategy text representation.}
We construct semantically rich and task-aware textual representations by integrating EOT, global-pooled, and attention-weighted token features, enabling more robust anomaly perception.

\textbf{Enhanced visual representation.}
We propose a multi-view foreground–background soft separation strategy along identity, semantic, and spatial dimensions, combined with background suppression, to capture fine-grained visual cues while reducing background noise.

\textbf{Cross-modal discriminative alignment.}
We introduce SCR to align image features with normal and abnormal textual prototypes, suppress uncertain matches, and enlarge similarity gaps, achieving stable cross-modal alignment.

%% file: sec/2_related.tex
\section{Related work}
\label{related_work}

\subsection{Prompt Design and Textual Adaptation}

In CLIP-based adaptation, prompt engineering plays a key role. Early methods such as CoOp~\cite{zhou2022learning} and CoCoOp~\cite{zhou2022conditional} learn context tokens for flexible textual adaptation, while AnomalyCLIP~\cite{zhou2023anomalyclip} and PromptAD~\cite{li2024promptad} use learnable prompts to capture generic or task-specific anomaly semantics. However, these approaches rely solely on EOT sentence embeddings, potentially overlooking token-level context.  
To address this, we propose \textbf{MSTFF}, which combines EOT, global pooling, and attention-based pooling to generate task-aware text representations that enhance anomaly sensitivity. Additionally, \textbf{SCR} enforces confidence and discriminative constraints on visual-text alignment, improving both text representations and ZSAD performance.

\subsection{Visual Feature Enhancement}

In terms of visual features, several works have attempted to enhance CLIP's spatial representations to improve anomaly detection. For example, WinCLIP~\cite{jeong2023winclip} achieves dense anomaly mapping via sliding-window patch evaluation, CLIP-AD~\cite{chen2024clip} utilizes local patch–text similarities for anomaly localization, VisionAD~\cite{taylor2024visionad} combines hierarchical features from multiple transformer layers, and AF-CLIP~\cite{fang2025af} employs neighborhood aggregation for feature enhancement. However, these methods primarily operate at the token level and overlook the semantic and structural information within each token. To address this, we propose the \textbf{MVFBE} and \textbf{BS} modules to disentangle and enhance semantic and spatial features, highlighting anomalous signals while suppressing background redundancy.

% \subsection{Visual Feature Enhancement and Foreground--Background Modeling}

% On the visual side, several works attempt to enhance CLIP's spatial representation for anomaly detection. WinCLIP~\cite{jeong2023winclip} applied sliding-window patch evaluation for dense anomaly mapping, and CLIP-AD~\cite{chen2024clip} utilized local patch–text similarities for localization. VisionAD~\cite{taylor2024visionad} combined hierarchical features from multiple transformer layers, while CLIP-SAM~\cite{hou2024enhancing} integrated CLIP~\cite{radford2021learning} with SAM~\cite{kirillov2023segment} segmentation priors to highlight object regions and suppress background noise. However, the segmentation in CLIP-SAM mainly relies on external masks rather than explicitly modeling the interaction between foreground and background at the feature level. 
% To address this, we propose the \textbf{MVFBE} and \textbf{BS} modules, which disentangle and enhance semantic and spatial features to highlight anomalies and suppress background redundancy.
% Our proposed \textbf{Multi-View Foreground–Background Enhancement (FEBG)} module addresses this limitation by disentangling and enhancing features from complementary semantic and spatial perspectives, ensuring that anomaly cues are emphasized while background redundancy is effectively suppressed.

%% file: sec/3_method.tex
\section{Method}
\label{3_method}

\begin{figure*}[htbp]
  \centering
  \includegraphics[width=0.95\textwidth]{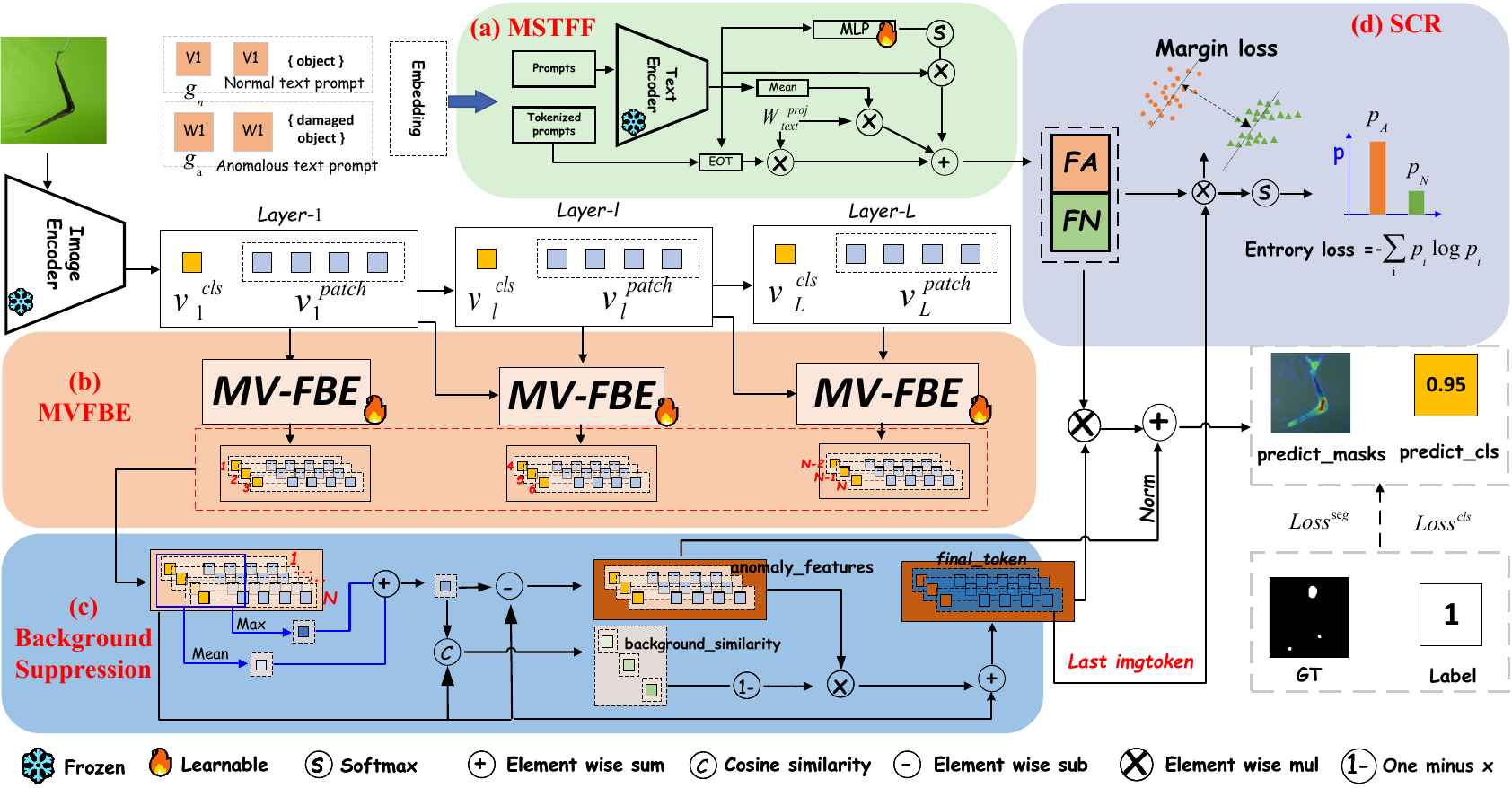}
  \caption{Overview of \textbf{FB-CLIP} for zero-shot anomaly detection. (a) Multi-Strategy Text Feature Fusion (MSTFF) generates task-aware text embeddings (see Section\ref{MSTFF}); (b) Multi-View Foreground-Background Enhancement (MVFBE) separates and enhances foreground and background features (see Section\ref{MVFBE}); (c) Background Suppression reduces residual background interference (see Section\ref{sec:BS}); (d) Semantic Consistency Regularization (SCR) enforces confident and discriminative visual-text alignment. These components jointly enable fine-grained anomaly detection without labeled abnormal samples (see Section\ref{SCR}).}
  \label{fig:FB-CLIP}
\end{figure*}

\subsection{Multi-Strategy Text Feature Fusion (MSTFF)}
\label{MSTFF}

Following \textbf{AnomalyCLIP}~\cite{zhou2023anomalyclip}, we adopt \textbf{object-agnostic learnable text prompts} for ZSAD, which discard class-specific semantics to capture general anomaly patterns. The normal and abnormal prompts are defined as:
\begin{align}
    g_n &= [V_1][V_2]\dots[V_E][\text{object}], \\
    g_a &= [W_1][W_2]\dots[W_E][\text{damaged}][\text{object}],
\end{align}
where $[V]_i$ and $[W]_i$ denote learnable embeddings for normal and abnormal prompts, respectively. This object-agnostic formulation directs the model to focus on anomaly-related cues rather than object categories, enabling better cross-object and cross-domain generalization.

CLIP’s text encoder typically relies on the End-of-Text (EOT) token for sentence-level representation, which may overlook contextual or anomaly-relevant semantics. To enhance textual representation, we propose a \textbf{Multi-Strategy Text Feature Fusion (MSTFF)} (as shown in Figure \ref{fig:FB-CLIP}(a)) module that combines complementary pooling strategies for richer and more discriminative text features .

Given the transformer-encoded token sequence $\mathbf{X} \in \mathbb{R}^{B \times L \times D}$, MSTFF extracts three types of features:\\
     \textbf{EOT feature} for maintaining CLIP compatibility:
    \begin{equation}
        \mathcal{F}_{\text{eot}} = 
        \mathbf{X}[\text{arange}(B), \text{argmax}(\mathbf{t}_{\text{token}})] \mathbf{W}_{\text{proj}}.
    \end{equation}
     \textbf{Global feature} via mean pooling to capture context:
    \begin{equation}
        \mathcal{F}_{\text{global}} = \frac{1}{L}\sum_{i=1}^{L}\mathbf{X}[:, i, :]\mathbf{W}_{\text{proj}}.
    \end{equation}
     \textbf{Attention feature} guided by a lightweight token selector $\mathcal{S}(\cdot)$ implemented as a two-layer MLP:
    \begin{align}
        % \alpha_i &= \text{softmax}\big(\mathcal{S}(\mathbf{X}[:, i, :])\big), \\
        \mathcal{F}_{\text{attn}} &= \sum_{i=1}^{L}\text{softmax}\big(\mathcal{S}(\mathbf{X}[:, i, :])\big) \mathbf{X}[:, i, :].
    \end{align}

Final text representation by weighted fusion:
\begin{equation}
    \mathcal{F}_{\text{text}} = 
    \lambda_1 \mathcal{F}_{\text{global}} +
    \lambda_2 \mathcal{F}_{\text{attn}}^{\text{proj}} +
    \lambda_3 \mathcal{F}_{\text{eot}},
\end{equation}
where $\lambda_1=1.0$, $\lambda_2=\lambda_3=0.5$.

Through this fusion, MSTFF captures complementary semantics: the EOT feature preserves CLIP alignment, the global feature provides contextual stability, and the attention feature highlights task-relevant tokens. This design yields expressive, adaptive, and generalizable text representations for zero-shot anomaly detection.

% ***********************************************************
% ***********************************************************
% ***********************************************************
\subsection{Multi-View Foreground-Background Enhancement (MVFBE)}
\label{MVFBE}

\begin{figure}[htbp]
  \centering
  \includegraphics[width=\columnwidth]{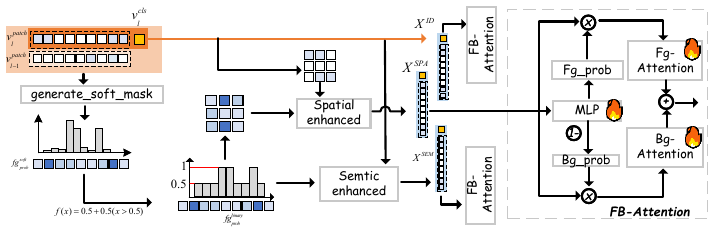}
  \caption{Structure of the proposed \textbf{MVFBE} module. The module first generates a soft foreground mask from multiple anomaly indicators. Three complementary views are applied: Identity (ID) preserves original features, Semantic (SEM) models foreground richness and background stability, and Spatial (SPA) captures fine-grained local structures. Features are then integrated across layers and refined with foreground-background attention to produce discriminative anomaly representations. Details in Appendix~\ref{soft_mask},~\ref{MVFBE_enhance}}.
    % \caption{Structure of the proposed \textbf{Foreground-Background Enhancement (FBE)} module. }
  \label{fig:FBE}
\end{figure}

% Existing ZSAD methods typically suffer from ambiguous foreground-background distinction, leading to degraded detection performance due to interference from redundant background information. To address this challenge, we propose a novel \textbf{Multi-View Foreground-Background Enhancement (MVFBE)} (as shown in Figure \ref{fig:FB-CLIP}(b)) module that explicitly separates and enhances foreground and background features from complementary perspectives. Our key insight is that foreground tokens, which typically contain anomalies, exhibit higher \textit{information richness} (diversity from the global context), while background tokens demonstrate greater \textit{stability} (consistency with normal patterns).

While MSTFF focuses on textual representations, visual features still suffer from foreground-background ambiguity. To address this challenge, we propose a novel \textbf{Multi-View Foreground-Background Enhancement (MVFBE)} (as shown in Figure \ref{fig:FB-CLIP}(b)) module that explicitly separates and enhances foreground and background features from complementary perspectives. Our key insight is that foreground tokens, which typically contain anomalies, exhibit higher \textit{information richness} (diversity from the global context), while background tokens demonstrate greater \textit{stability} (consistency with normal patterns).

Given input visual tokens $\mathbf{X} \in \mathbb{R}^{B \times (L+1) \times C}$ extracted from a pre-trained Vision Transformer (ViT), including one class token $\mathbf{x}_{\text{cls}} \in \mathbb{R}^{B \times 1 \times C}$ and $L$ patch tokens $\mathbf{x}_{\text{img}} \in \mathbb{R}^{B \times L \times C}$, the MVFBE module first generates a soft foreground probability mask $\mathbf{P}_{\text{fg}} \in \{0.5, 1.0\}^{B \times L}$ using multiple complementary anomaly indicators. Specifically, we compute four complementary cues: (1) \textit{local saliency} $\mathcal{S}_{\text{local}}$  (2) \textit{center distance} $\mathcal{D}_{\text{center}}$ (3) \textit{CLS inconsistency} $\mathcal{I}_{\text{cls}}$  and (4) \textit{temporal variation} $\mathcal{T}_{\text{temp}}$. Details in Appendix~\ref{soft_mask}.

% To ensure fair contribution from each indicator and prevent inter-sample information leakage, we apply per-sample min-max normalization to each indicator:
% \begin{equation}
% \tilde{\mathcal{S}}_i = \frac{\mathcal{S}_i - \min_{j}(\mathcal{S}_i[j])}{\max_{j}(\mathcal{S}_i[j]) - \min_{j}(\mathcal{S}_i[j]) + \epsilon},
% \end{equation}
% where $\mathcal{S}_i \in \{\mathcal{S}_{\text{local}}, \mathcal{D}_{\text{center}}, \mathcal{I}_{\text{cls}}, \mathcal{T}_{\text{temp}}\}$, $j$ indexes tokens within each sample, and $\epsilon = 10^{-6}$ prevents numerical instability. 

The normalized indicators are then linearly combined to form an anomaly score:
\begin{equation}
 \mathcal{A} = \alpha_1 \tilde{\mathcal{S}}_{\text{local}} + \alpha_2 \tilde{\mathcal{D}}_{\text{center}} + \alpha_3 \tilde{\mathcal{I}}_{\text{cls}} + \alpha_4 \tilde{\mathcal{T}}_{\text{temp}},
\end{equation}
where $\alpha_1 = \alpha_2 = \alpha_3 = 0.3$ and $\alpha_4 = 0.1$ to emphasize spatial and semantic cues while incorporating temporal information when available. 
The anomaly score is then binarized to generate the soft foreground mask:
\begin{equation}
\mathbf{P}_{\text{fg}}[i] = \begin{cases}
1.0 & \text{if } \mathcal{A}[i] > 0.5 \\
0.5 & \text{otherwise}
\end{cases},
\end{equation}
where the threshold of $0.5$ is chosen based on the per-sample normalized score distribution. The soft mask design with values $\{0.5, 1.0\}$ maintains gradient stability during backpropagation while ensuring clear separation: tokens with $\mathbf{P}_{\text{fg}}[i] = 1.0$ represent high-confidence foreground regions, while those with $\mathbf{P}_{\text{fg}}[i] = 0.5$ are treated as uncertain/background tokens that require further refinement in subsequent enhancement stages.

\subsubsection{Three Complementary Enhancement Views}

The MVFBE module applies three complementary enhancement views to process the visual tokens.

\textbf{Identity View (ID).}
The \textbf{Identity (ID)} view preserves original features:
\begin{equation}
\mathbf{X}_{\text{ID}} = \mathbf{X}.
\end{equation}
This serves as a baseline to retain raw feature information and prevent degradation from over-transformation.

\textbf{Semantic View (SEM).}
The \textbf{SEM} view performs global self-attention-based enhancement by explicitly modeling \textit{information richness} for foreground tokens and \textit{stability} for background tokens. (The pseudocode in Appendix~\ref{code_sem})

We first construct interaction weight matrices:
\begin{equation}
\mathbf{W}_{\text{fg}} = \mathbf{p}_{\text{fg}} \otimes \mathbf{p}_{\text{fg}}^T, \quad
\mathbf{W}_{\text{bg}} = (1 - \mathbf{p}_{\text{fg}}) \otimes (1 - \mathbf{p}_{\text{fg}})^T,
\end{equation}
where $\otimes$ denotes outer product.  
Foreground information richness is computed by deviation from the class token:
\begin{equation}
\mathbf{r}_{\text{info}} = 1 - \text{cosine}(\mathbf{X}_{\text{tokens}}, \mathbf{X}_{\text{cls}}),
\end{equation}
and attention weights are:
\begin{equation}
\mathbf{A}_{\text{fg}} = \text{Softmax}(\mathbf{r}_{\text{info}} \odot \mathbf{W}_{\text{fg}}).
\end{equation}
For background tokens, stability is measured by similarity:
\begin{equation}
\begin{aligned}
\mathbf{s}_{\text{stable}} &= \text{cosine}(\mathbf{X}_{\text{tokens}}, \mathbf{X}_{\text{cls}}), \\
\mathbf{A}_{\text{bg}} &= \text{Softmax}(\mathbf{s}_{\text{stable}} \odot \mathbf{W}_{\text{bg}}).
\end{aligned}
\end{equation}
Foreground and background features are aggregated:
\begin{equation}
\mathbf{X}_{\text{fg,agg}} = \mathbf{A}_{\text{fg}}\mathbf{X}_{\text{tokens}}, \quad
\mathbf{X}_{\text{bg,agg}} = \mathbf{A}_{\text{bg}}\mathbf{X}_{\text{tokens}},
\end{equation}
and fused with a residual connection:
\begin{equation}
\begin{aligned}
\mathbf{X}_{\text{SEM}} = \alpha [&\mathbf{p}_{\text{fg}} \odot \mathbf{X}_{\text{fg,agg}} 
+ (1 - \mathbf{p}_{\text{fg}}) \odot \mathbf{X}_{\text{bg,agg}}] \\
&+ (1 - \alpha)\mathbf{X}_{\text{tokens}},  \alpha = 0.6.
\end{aligned}
\end{equation}

\textbf {Spatial View (SPA).}
Inspired by PatchCore~\cite{roth2022towards}, MUSC~\cite{li2024musc} , AF-CLIP~\cite{fang2025af}, CLIP-Haze~\cite{zhang2024adapt}, and MFP-CLIP~\cite{yuan2025mfp}, we design a spatial enhancement module from the perspective of neighborhood aggregation to more effectively capture local contextual information. 
Unlike existing methods, we explicitly utilize the unfolded mask to distinguish foreground and background patches, and further process them based on foreground information richness, introducing a fundamentally different spatial modeling paradigm.

The \textbf{Spatial (SPA)} view performs local patch-based enhancement to capture fine-grained spatial structures. Tokens are reshaped into 2D grids and unfolded into $5 \times 5$ overlapping patches (The pseudocode in Appendix~\ref{code_spa}):
\begin{equation}
\mathbf{X}_{\text{unfold}} \in \mathbb{R}^{(BHW) \times 25 \times C}.
\end{equation}
Foreground and background patches are separated using the unfolded mask.  
Background stability combines cosine similarity with feature magnitude:
\begin{equation}
\mathbf{s}_{\text{bg}} = \text{cosine}(\mathbf{X}_{\text{bg,patches}}, \mathbf{X}_{\text{cls}}) \odot \|\mathbf{X}_{\text{bg,patches}}\|,
\end{equation}
while foreground information richness considers deviation and local diversity:
\begin{equation}
\begin{aligned}
\mathbf{r}_{\text{fg}} = &(1 - \text{cosine}(\mathbf{X}_{\text{fg,patches}}, \mathbf{X}_{\text{cls}})) \\
&\odot \|\mathbf{X}_{\text{fg,patches}} - \mu(\mathbf{X}_{\text{fg,patches}})\|.
\end{aligned}
\end{equation}

Weighted aggregation produces the enhanced features:
% \begin{align}
% \mathbf{X}_{\text{bg,agg}} &= 
% \frac{\sum_{i=1}^{25} \text{Softmax}(\mathbf{s}_{\text{bg}})[i] \odot \mathbf{X}_{\text{bg,patches}}[i]}{\sum_{i=1}^{25} \mathbf{P}_{\text{bg,unfold}}[i]}, \\
% \mathbf{X}_{\text{fg,agg}} &= 
% \frac{\sum_{i=1}^{25} \text{Softmax}(\mathbf{r}_{\text{fg}})[i] \odot \mathbf{X}_{\text{fg,patches}}[i]}{\sum_{i=1}^{25} \mathbf{P}_{\text{fg,unfold}}[i]}, \\
% \mathbf{X}_{\text{SPA}} &= \text{Reshape}(\mathbf{X}_{\text{fg,agg}} + \mathbf{X}_{\text{bg,agg}}).
% \end{align}
\begin{align}
\mathbf{X}_{\text{fg/bg,agg}} &= 
\frac{\sum_{i=1}^{25} \text{Softmax}(\mathbf{w}_{\text{fg/bg}})[i] \odot \mathbf{X}_{\text{fg/bg,patches}}[i]}{\sum_{i=1}^{25} \mathbf{P}_{\text{fg/bg,unfold}}[i]}, \\
\mathbf{X}_{\text{SPA}} &= \text{Reshape}(\mathbf{X}_{\text{fg,agg}} + \mathbf{X}_{\text{bg,agg}}).
\end{align}

\subsubsection{Multi-Layer Integration}

The three enhanced views form:
\begin{equation}
\mathbf{F}_{\text{enhanced}} = [\mathbf{X}_{\text{ID}}, \mathbf{X}_{\text{SEM}}, \mathbf{X}_{\text{SPA}}],
\end{equation}
and are applied across $N$ Transformer layers, producing $3N$ enhanced feature groups.
Inspired by AF-CLIP~\cite{fang2025af}, each feature group is further processed by the \textbf{Foreground-Background Attention (FB-Attention)} (shown in Fig \ref{fig:FBE}) module with adaptive gating (inspired by LSTM~\cite{hochreiter1997long, hu2024specslice} ):
\begin{equation}
\mathbf{x}_{\text{fg},i} = \mathcal{G}(\mathbf{x}_i)\mathbf{x}_i, \quad
\mathbf{x}_{\text{bg},i} = (1 - \mathcal{G}(\mathbf{x}_i))\mathbf{x}_i,
\end{equation}
where $\mathcal{G}: \mathbb{R}^C \rightarrow [0,1]$ is a learnable gate function.  
Apply multi-head self-attention (MHSA)~\cite{vaswani2017attention} to each stream separately, then sum the outputs and apply LayerNorm (LN).
\begin{equation}
\mathbf{H}_{\text{FB}} = \text{LN}(\text{MHSA}_{\text{fg}}(\mathbf{X}_{\text{fg}}) + \text{MHSA}_{\text{bg}}(\mathbf{X}_{\text{bg}})).
\end{equation}

% ***********************************************************
% ***********************************************************
% ***********************************************************

\subsection{Background Suppression (BS)}
\label{sec:BS}

After MVFBE, we address the critical challenge of \textbf{background interference} in ZSAD. 
Although the enhancement module separates semantic regions, background tokens may still retain residual noise that degrades detection accuracy. 
To alleviate this, we introduce a \textbf{background subtraction mechanism} (as shown in Figure \ref{fig:FB-CLIP}(c)) that explicitly models and suppresses background patterns while preserving anomaly-specific signals.

% Given the $3N$ enhanced features $\mathbf{F}_{\text{enhanced}}$ from the MVFBE module, 
% we first extract a prototypical background representation by aggregating candidate background tokens across all features. 
% Following the observation that early-positioned tokens tend to encode background information, 
% we collect the first half of tokens from each enhanced feature:
Given the $3N$ enhanced features $\mathbf{F}_{\text{enhanced}}$, we extract a prototypical background representation by aggregating the first half of candidate background tokens from each feature.
\begin{equation}
\mathbf{X}_{\text{bg,cand}}^{(i)} 
= \mathbf{F}_{\text{enhanced}}^{(i)}[:, :L/2, :].
\end{equation}

These candidates are concatenated to form a unified background feature bank:
\begin{equation}
\mathbf{X}_{\text{bg,bank}} 
= \text{Concat}\big[\mathbf{X}_{\text{bg,cand}}^{(1)}, 
\ldots, \mathbf{X}_{\text{bg,cand}}^{(3N)}\big],
\end{equation}
where $\mathbf{X}_{\text{bg,bank}} \in \mathbb{R}^{B \times M \times C}$ and $M$ is the total number of background tokens. 
The prototypical background vector is computed as:
\begin{equation}
\mathbf{b}_{\text{proto}} 
= \tfrac{1}{2}\,\text{Mean}(\mathbf{X}_{\text{bg,bank}}) 
+ \tfrac{1}{2}\,\text{Max}(\mathbf{X}_{\text{bg,bank}}),
\end{equation}
which combines global averaging and max-pooling to preserve salient background characteristics.

For each feature $\mathbf{F}_{\text{enhanced}}^{(i)} \in \mathbb{R}^{B \times L \times C}$, 
we compute cosine similarity to the background prototype:
\begin{equation}
\mathbf{s}_{\text{bg}}^{(i)} 
= \text{cosine}\!\left(\mathbf{F}_{\text{enhanced}}^{(i)}, 
\mathbf{b}_{\text{proto}}\right)
\in \mathbb{R}^{B \times L \times 1}.
\end{equation}

The anomaly signal is then obtained by background subtraction:
\begin{equation}
\mathbf{a}^{(i)} 
= \mathbf{F}_{\text{enhanced}}^{(i)} - \mathbf{b}_{\text{proto}},
\end{equation}
and its reconstruction error is defined as:
\begin{equation}
\mathbf{e}^{(i)} = \|\mathbf{a}^{(i)}\|_2.
\end{equation}

To suppress residual background while emphasizing anomaly cues, 
we apply similarity-weighted enhancement:
\begin{equation}
\mathbf{a}_{\text{enh}}^{(i)} 
= \mathbf{a}^{(i)} \odot (1 - \mathbf{s}_{\text{bg}}^{(i)}),
\end{equation}
where tokens highly similar to the background receive smaller weights. 
The final background-suppressed features:
\begin{equation}
\mathbf{F}_{\text{final}}^{(i)} 
= \alpha\,\mathbf{F}_{\text{enhanced}}^{(i)} 
+ (1-\alpha)\,\mathbf{a}_{\text{enh}}^{(i)}, 
\quad \alpha = 0.5.
\end{equation}

\subsection{Semantic Consistency Regularization (SCR)}
\label{SCR}

Even after visual disentanglement, visual-text alignment may remain unstable in the zero-shot setting. Therefore, inspired by Support Vector Machines~\cite{hearst1998support},
we introduce a \textbf{semantic consistency regularization} (as shown in Figure \ref{fig:FB-CLIP}(d)) loss that jointly enforces \textit{confidence} and \textit{discriminative} constraints on visual--text alignment without additional supervision.
Given image token $\mathbf{V}_{\text{tokens}}\!\in\!\mathbb{R}^{B\times N\times D}$ (from the last layer after background suppression, representing the most semantically rich anomaly-focused representations) and text features $\mathbf{T}\!\in\!\mathbb{R}^{2\times D}$ (for ``normal'' and ``abnormal'' prompts), 
we aggregate token-level features via global average pooling and apply L2 normalization:
\begin{equation}
% \mathbf{V} = \tfrac{1}{N}\sum_{n=1}^{N} \mathbf{V}_{\text{tokens}}[:,n,:], \quad
\mathbf{V}\leftarrow\text{L2Norm}(\mathbf{V}), \quad 
\mathbf{T}\leftarrow\text{L2Norm}(\mathbf{T}).
\end{equation}
Semantic similarity is computed as:
\begin{equation}
\mathbf{s} = \tfrac{1}{\tau}\mathbf{V}\mathbf{T}^{\top}, \quad 
\mathbf{s}\in\mathbb{R}^{B\times2}, \ \tau=0.07,
\end{equation}
where $\mathbf{s}[b,0]$ and $\mathbf{s}[b,1]$ denote similarities to normal and abnormal prompts, respectively.
Class probabilities are:
\begin{equation}
\mathbf{p}_b = \text{Softmax}(\mathbf{s}_b), \quad \mathbf{p}_b\in\mathbb{R}^2,
\end{equation}
% where $p_b[0]$ and $p_b[1]$ represent probabilities for normal and abnormal classes.

\paragraph{Entropy Regularization.}
We encourage confident predictions by minimizing entropy:
\begin{equation}
% \small
\mathcal{L}_{\text{entropy}}
= \mathbb{E}_{b=1}^{B}
\!\left[ -\sum_{c=0}^{1} p_b(c)\log\left(p_b(c) \right) \right]
\end{equation}
% \begin{equation}
% \resizebox{0.8\linewidth}{!}{%
% $\mathcal{L}_{\text{entropy}} = 
% \frac{1}{B} \sum_{b=1}^{B} \Big[-\sum_{c=0}^{1} p_b(c)\log(p_b(c)+\epsilon)\Big],\ \epsilon = 10^{-8}$%
% }
% \end{equation}

% where $\epsilon = 10^{-8}$ ensures numerical stability.

\paragraph{Margin Regularization.}
We enforce discriminative separation via margin constraint:
\begin{equation}
\mathcal{L}_{\text{margin}}
= \mathbb{E}_{b=1}^{B}
\!\left[ \max\!\left(0,\, \gamma - |s_b[1] - s_b[0]| \right) \right],
\end{equation}
where $\gamma = 1$ is the margin parameter. This prevents mode collapse and maintains clear separation between normal and abnormal semantics.
% \paragraph{Combined Objective.}
% The final loss combines both terms:
% \begin{equation}
% \mathcal{L}_{\text{consistency}}
% = \lambda \big( 
% w_e\,\mathcal{L}_{\text{entropy}} 
% + w_m\,\mathcal{L}_{\text{margin}} 
% \big),
% \end{equation}
% where $\lambda = 0.15$, $w_e = 1.0$, and $w_m = 0.5$.
% This dual regularization simultaneously promotes confident predictions and maintains discriminative separation, enhancing the model's ability to leverage pre-trained CLIP representations in zero-shot scenarios.
The final loss combines both terms:
\begin{equation}
\mathcal{L}_{\text{consistency}}
= \lambda \big( 
w_e\,\mathcal{L}_{\text{entropy}} 
+ w_m\,\mathcal{L}_{\text{margin}} 
\big),
\end{equation}
where $\lambda = 0.15$ controls the overall weight, and $w_e = 1.0$, $w_m = 0.5$ 
are relative coefficients for the entropy and margin terms.
This dual regularization strategy addresses two essential aspects :
(1) encouraging confident and semantically consistent predictions, 
(2) maintaining discriminative separation between normal and abnormal semantics.  

%% file: sec/4_exe.tex
\section{Experiment}
\label{exe}

\subsection{Experiment setup}

\begin{table*}[h]
\centering
\caption{Zero-shot anomaly detection performance on industrial and medical domains. The image-level performance AUROC(\%), AP(\%) and pixel-level performance AUROC(\%), PRO(\%) are reported. \textcolor{red}{\textbf{Red: best}}, \textcolor{blue}{\textbf{blue: second best.}}}
\resizebox{\textwidth}{!}{
\begin{tabular}{ccccccccccccccc}
\toprule
\multirow{2}{*}{Domain} & \multirow{2}{*}{Metric} & \multirow{2}{*}{Dataset} 
& WinCLIP~\cite{jeong2023winclip} & VAND~\cite{chen2023zero} & AnomalyCLIP~\cite{zhou2023anomalyclip} & AdaCLIP~\cite{cao2024adaclip} & AA-CLIP~\cite{ma2025aa} & AF-CLIP~\cite{fang2025af} & FAPrompt~\cite{zhu2025fine} & FB-CLIP\\
& & & CVPR'2023 & CVPR'2023 & ICLR'2024 & ECCV'2024 & CVPR'2025 &  MM'2025 & ICCV'2025 & Ours\\
\midrule
\multirow{12}{*}{Industrial} 
& \multirow{5}{*}{\shortstack{image-level \\ (AUROC, AP)}} 
& MVTec & (91.8, 96.5) & (86.1, 93.5) & (91.5, 96.2) & (92.1, \textcolor{blue}{\textbf{96.7}}) & (90.5, 94.9) & (\textcolor{red}{\textbf{92.9}}, \textcolor{red}{\textbf{96.8}}) & (91.9, 95.7) & (\textcolor{blue}{\textbf{92.4}},96.6)\\
& & VisA & (78.1, 81.2) & (78.0, 81.4) & (82.1, 85.4) & (86.3, 89.2) & (84.6, 82.2) & (\textcolor{blue}{\textbf{88.5}}, \textcolor{blue}{\textbf{90.0}}) & (84.6, 86.8) & (\textcolor{red}{\textbf{89.5}}, \textcolor{red}{\textbf{90.7}})\\
& & MPDD & (63.6, 69.9) & (78.0, 81.4) & (77.0, 82.0) & (76.8, -) & (-, -) & (-, -) & (\textcolor{red}{\textbf{80.1}}, \textcolor{blue}{\textbf{83.9}}) & (\textcolor{blue}{\textbf{79.1}}, \textcolor{red}{\textbf{86.2}})\\
& & BTAD & (68.2, 70.9) & (73.5, 68.6) & (88.3, 87.3) & (92.4, 96.2) & (\textcolor{blue}{\textbf{93.8}}, \textcolor{red}{\textbf{97.9}}) & (\textcolor{red}{\textbf{94.3}}, 95.2) & (92.2, 92.5) & (93.2, \textcolor{blue}{\textbf{96.4}})\\
& & DAGM & (91.8, 79.5) & (94.4, 83.8) & (97.5, 92.3) & (98.3, 93.7) & (93.9, 84.5) & (98.7, 94.6) & (\textcolor{blue}{\textbf{98.8}}, \textcolor{blue}{\textbf{95.3}}) & (\textcolor{red}{\textbf{99.0}}, \textcolor{red}{\textbf{96.5}})\\
& & DTD & (93.2, 92.6) & (86.4, 95.0) & (93.5, 97.0) & (97.0, 99.0) & (93.3, 97.8) & (\textcolor{red}{\textbf{97.9}}, \textcolor{red}{\textbf{99.1}}) & (96.2, 98.1) & (\textcolor{red}{\textbf{97.9}}, \textcolor{red}{\textbf{99.1}})\\

& & Real-IAD & (-, -) & (-, -) &  (-, -) & (-, -) & (-, -) &(\textcolor{blue}{\textbf{79.2}}, \textcolor{blue}{\textbf{77.0}}) & (77.3, 74.8) & (\textcolor{red}{\textbf{80.6}}, \textcolor{red}{\textbf{78.4}})\\

\cmidrule(lr){2-11}
& \multirow{5}{*}{\shortstack{pixel-level \\ (AUROC, PRO)}} 
& MVTec & (85.1, 64.6) & (87.6, 44.0) & (91.1, 81.4) & (86.3, 20.0) & (\textcolor{blue}{\textbf{91.9}}, 84.6) & (\textcolor{red}{\textbf{92.3}}, \textcolor{red}{\textbf{85.7}}) & (90.6, 83.3) & (\textcolor{blue}{\textbf{91.9}}, \textcolor{red}{\textbf{85.7}})\\
& & VisA & (79.6, 56.8) & (94.2, 86.8) & (95.5, 87.0) & (95.8, 60.0) & (95.5, 83.0) & (\textcolor{blue}{\textbf{96.2}}, \textcolor{blue}{\textbf{88.7}}) & (95.9, 87.7) & (\textcolor{red}{\textbf{96.3}}, \textcolor{red}{\textbf{91.4}})\\
& & MPDD & (76.4, 48.9) & (94.1, 83.2) & (\textcolor{blue}{\textbf{96.5}}, 87.0) & (96.1, -) & (-, -) & (-, -) & (\textcolor{blue}{\textbf{96.5}}, \textcolor{blue}{\textbf{87.9}}) & (\textcolor{red}{\textbf{96.9}}, \textcolor{red}{\textbf{91.1}})\\
& & BTAD & (72.7, 27.3) & (89.3, 68.8) & (94.2, 74.8) & (94.0, 33.8) & (94.0, 69.0) & (94.4, \textcolor{blue}{\textbf{78.3}}) & (\textcolor{blue}{\textbf{95.6}}, 75.1) & (\textcolor{red}{\textbf{95.8}}, \textcolor{red}{\textbf{80.0}})\\
& & DAGM & (87.6, 65.7) & (82.4, 66.2) & (95.6, 91.0) & (94.5, 53.8) & (91.6, 76.5) & (97.1, 93.1) & (\textcolor{red}{\textbf{98.2}}, \textcolor{red}{\textbf{95.0}}) & (\textcolor{blue}{\textbf{97.3}}, \textcolor{blue}{\textbf{94.1}})\\
& & DTD & (83.9, 57.8) & (95.3, 86.9) & (97.9, 92.3) & (98.5, 72.9) & (96.4, 85.9) & (\textcolor{red}{\textbf{98.6}}, \textcolor{blue}{\textbf{93.8}}) & (\textcolor{blue}{\textbf{98.3}}, \textcolor{blue}{\textbf{93.3}}) & (\textcolor{blue}{\textbf{98.3}}, 92.6)\\
& & Real-IAD & (-, -) & (-, -) &  (-, -) & (-, -) & (-, -) &  (\textcolor{blue}{\textbf{95.5}}, 81.6) & (95.0, \textcolor{blue}{\textbf{82.1}}) & (\textcolor{red}{\textbf{95.9}}, \textcolor{red}{\textbf{88.2}})\\
\midrule
\multirow{9}{*}{Medical} 
& \multirow{3}{*}{\shortstack{image-level \\ (AUROC, AP)}} 
& HeadCT & (81.8, 80.2) & (89.1, 89.4) & (\textcolor{blue}{\textbf{93.4}}, 91.6) & (91.4, -) & (-, -) &  (91.2, 91.1) & (\textcolor{red}{\textbf{94.0}}, \textcolor{blue}{\textbf{92.4}}) & (\textcolor{blue}{\textbf{93.4}}, \textcolor{red}{\textbf{93.9}})
\\
& &BrainMRI & (86.6, 91.5) & (89.3, 90.9) & (90.3, 92.3) & (80.6, 84.4) & (91.8, 94.1) & (\textcolor{blue}{\textbf{95.2}}, \textcolor{blue}{\textbf{96.3}}) & (\textcolor{red}{\textbf{95.8}}, 96.2) & (95.1, \textcolor{red}{\textbf{96.6}})\\
& & Br35H & (80.5, 82.2) & (93.1, 92.9) & (94.6, 94.7) & (86.5, 86.3) & (89.4, 91.0) & (96.7, 96.4) & (\textcolor{red}{\textbf{97.6}}, \textcolor{blue}{\textbf{97.1}}) & (\textcolor{blue}{\textbf{97.1}}, \textcolor{red}{\textbf{97.4}})\\
\cmidrule(lr){2-11}
& \multirow{6}{*}{\shortstack{pixel-level \\ (AUROC, PRO)}} 
& ISIC & (83.3, 55.1) & (89.4, 77.2) & (89.7, 78.4) & (89.3, -) & (\textcolor{blue}{\textbf{93.9}}, 87.0) & (\textcolor{red}{\textbf{94.8}},\textcolor{red}{\textbf{89.6}}) & (91.1, 81.6) & (\textcolor{blue}{\textbf{93.9}}, \textcolor{blue}{\textbf{87.7}})\\
& & ClinicDB & (51.2, 13.8) & (80.5, 60.7) & (82.9, 67.8) & (81.6, 54.0) & (86.3, 67.1) & (\textcolor{blue}{\textbf{87.1}}, 70.0) & (84.7, \textcolor{blue}{\textbf{70.1}}) & (\textcolor{red}{\textbf{87.2}}, \textcolor{red}{\textbf{73.5}})\\
&  &ColonDB & (70.3, 32.5) & (78.4, 64.6) & (81.9, 71.3) & (79.3, 49.1) & (81.6, 65.4) & (83.2, 67.9) & (\textcolor{red}{\textbf{85.0}}, \textcolor{blue}{\textbf{73.3}}) & (\textcolor{blue}{\textbf{84.2}}, \textcolor{red}{\textbf{74.4}})\\
& & Kvasir & (69.7, 24.5) & (75.0, 36.2) & (78.9, 45.6) &  (77.5, 41.3) & (84.0, \textcolor{blue}{\textbf{50.3}}) & (\textcolor{blue}{\textbf{84.5}}, \textcolor{red}{\textbf{57.5}}) &(82.1, 49.9) & (\textcolor{red}{\textbf{85.0}}, 47.8)\\
& & Endo & (68.2, 28.3) & (81.9, 54.9) & (84.1, 63.6) & (-, -) & (-, -) & (\textcolor{red}{\textbf{88.3}}, \textcolor{blue}{\textbf{70.3}}) &  (86.8, 67.6) & (\textcolor{blue}{\textbf{87.6}}, \textcolor{red}{\textbf{70.4}})\\
& & TN3K & (70.7, 39.8) & (73.6, 37.8) & (\textcolor{blue}{\textbf{81.5}}, \textcolor{blue}{\textbf{50.4}}) & (77.2, -) & (-, -) &  (80.5, 45.8) &(\textcolor{red}{\textbf{84.7}}, \textcolor{red}{\textbf{54.6}}) &(79.0, 46.8)\\
\bottomrule
\end{tabular}
}
\label{tab:zero-shot-colored}
\end{table*}

\textbf{Datasets and Evaluation Metrics.}
We evaluate FB-CLIP on 16 publicly available datasets spanning industrial inspection and medical imaging. Industrial benchmarks include MVTec AD~\citep{bergmann2019mvtec}, VisA~\citep{zou2022spot}, MPDD~\citep{jezek2021deep}, BTAD~\citep{mishra2021vt}, DAGM~\citep{wieler2007weakly}, DTD-Synthetic~\citep{aota2023zero} and a large-scale dataset Real-IAD~\cite{wang2024real}, while medical datasets cover colon polyp (CVC-ClinicDB~\citep{bernal2015wm}, CVC-ColonDB~\citep{tajbakhsh2015automated}), skin cancer detection dataset ISIC~\citep{gutman2016skin} and brain tumor (BrainMRI~\citep{salehi2021multiresolution}, Br35H~\citep{br35h}) detection,  Kvasir~\citep{jha2020kvasir}, and Endo~\citep{hicks2021endotect}, thyroid nodule detection dataset TN3k~\citep{gong2021multi}, brain tumor detection datasets HeadCT~\citep{salehi2021multiresolution}. 
Competing SOTA methods include WinCLIP~\citep{jeong2023winclip}, VAND~\citep{chen2023zero}, AnomalyCLIP~\citep{zhou2023anomalyclip}, AdaCLIP~\citep{cao2024adaclip}, AA-CLIP~\citep{ma2025aa}, AF-CLIP~\citep{fang2025af}, and FAPrompt~\citep{zhu2025fine}. Performance is measured using AUROC for detection, AP for anomaly classification, and AUPRO~\citep{bergmann2020uninformed} for localization.

% \paragraph{Implementation Details}
\textbf{Implementation Details.}
\label{Implementation details}
FB-CLIP is built on the public CLIP backbone (\verb|ViT-L/14@336px|) with frozen parameters. Following prior works~\citep{jeong2023winclip,zhou2023anomalyclip,fang2025af}, we fine-tune AnomalyCLIP using the test data on MVTec AD and
evaluate the ZSAD performance on other datasets. As for MVTec AD, we fine-tune AomalyCLIP
on the test data of VisA. Results are averaged over all sub-datasets. All experiments are implemented in PyTorch~2.0.0 and conducted on a single NVIDIA RTX 3090 GPU (24GB) using the Adam optimizer with a learning rate of 5e-5 and a batch size of 4.

\subsection{Main results}

As shown in Table \ref{tab:zero-shot-colored}, FB-CLIP demonstrates competitive performance in zero-shot anomaly detection across both industrial and medical domains. Compared with recent strong CLIP-based models such as AF-CLIP and FAPrompt, our method consistently achieves superior results on both image-level and pixel-level metrics in the industrial domain. FB-CLIP achieves notable improvements on the VisA dataset and attains over a 6\% increase in pixel-level AUPRO on the large-scale Real-IAD dataset. While FAPrompt remains competitive on certain medical datasets, FB-CLIP ranks among the top two in 14 out of 18 metrics, surpassing FAPrompt, which ranks in the top two on 10 metrics.

% FB-CLIP achieves new state-of-the-art performance in zero-shot anomaly detection across both industrial and medical domains. Compared with recent strong CLIP-based models such as AF-CLIP and FAPrompt , our method consistently attains superior results at both image- and pixel-level metrics. In the industrial domain, FB-CLIP outperforms AF-CLIP on almost all datasets, achieving notable gains on VisA (+0.8 AUROC / +0.7 AP) and BTAD (+1.4 PRO). While FAPrompt performs competitively on certain datasets, FB-CLIP surpasses it on 9 out of 12 metrics, establishing the best localization performance on MVTec, VisA, and MPDD. In the medical domain, FB-CLIP further improves over AF-CLIP and FAPrompt, achieving the highest pixel-level AUROC/PRO scores on ClinicDB (87.2 / 73.5) and ColonDB (84.2 / 74.4).

\subsection{Ablation Study}

\begin{table}[htbp]
    \centering
    \caption{Ablation study of different modules for zero-shot anomaly detection. Deeper colors correspond to higher average performance. The columns \textbf{MSTFF}, \textbf{MVFBE}, \textbf{BS}, and \textbf{SCR} correspond to the four improvement strategies proposed in Figure~\ref{fig:FB-CLIP}.}
    \setlength{\tabcolsep}{5pt}
    \renewcommand{\arraystretch}{1.25}
    \resizebox{\columnwidth}{!}{
    \begin{tabular}{
        >{\centering\arraybackslash}p{1.3cm}
        >{\centering\arraybackslash}p{1.3cm}
        >{\centering\arraybackslash}p{1.3cm}
        >{\centering\arraybackslash}p{1.3cm}
        >{\centering\arraybackslash}p{1.3cm}
        cccc
    }
    \toprule
        \multicolumn{5}{c}{\textbf{Method}} 
        & \multicolumn{2}{c}{\textbf{MVTec}} 
        & \multicolumn{2}{c}{\textbf{VisA}} \\
        \cmidrule(lr){1-5} \cmidrule(lr){6-7} \cmidrule(lr){8-9}
        \textbf{Base} & \textbf{MSTFF} & \textbf{MVFBE} & \textbf{BS} & \textbf{SCR} 
        & \shortstack{\textbf{Image} \\ (AUROC, AP)} 
        & \shortstack{\textbf{Pixel} \\ (AUROC, PRO)} 
        & \shortstack{\textbf{Image} \\ (AUROC, AP)} 
        & \shortstack{\textbf{Pixel} \\ (AUROC, PRO)} \\
    \midrule
        \rowcolor{gray!5}
        \cmark &  &  &  &  &   (67.1, 84.0) & (65.2, 32.9) & (45.7, 56.2) & (88.4, 71.8) \\

        \cmark & \cmark &  &  &  & (82.0, 91.4) & (84.0, 68.0) & (78.3, 80.5) & (93.2, 81.4) \\

        \cmark &  & \cmark &  &  & (92.2, 96.5) & (91.7, 84.5) & (88.4, 89.7) & (96.0, 90.8) \\

        \cmark &  &  & \cmark &  & (44.8, 72.7) & (70.7, 38.3) & (48.3, 56.6) & (84.1, 58.3) \\

        \cmark &  &  &  & \cmark & (65.4, 81.3) & (75.7, 45.5) & (63.8, 69.0) & (88.2, 65.5) \\

        \hdashline
        \rowcolor{gray!15}
        \cmark & \cmark & \cmark &  &  & (92.3, 96.5) & (91.2, 84.7) & (88.2, 89.7) & (95.9, 90.6) \\

        \cmark & \cmark &  & \cmark &  & (73.0, 86.8) & (71.2, 28.7) & (68.3, 72.2) & (86.5, 60.2) \\

        \rowcolor{gray!15}
        \cmark & \cmark &  &  & \cmark & (78.2, 88.8) & (83.2, 63.6) & (74.0, 74.6) & (92.4, 78.8) \\

        \cmark & \cmark & \cmark & \cmark &  & \textbf{(92.8, 96.8)} & (91.6, \textbf{85.7}) & (87.6, 89.3) & (96.1, 90.6) \\
        
        \rowcolor{gray!15}
        \cmark & \cmark & \cmark &  & \cmark & (92.1, 96.2) & (89.4, 78.2) & (87.9, 89.8) & (96.2, 90.7) \\
   
        \cmark &  & \cmark & \cmark &  & (92.4, 96.7) & (91.4, 85.6) & (87.8, 89.4) & (96.1, 90.7) \\

        \rowcolor{gray!15}
        \cmark &  & \cmark &  & \cmark & (92.1, 96.3) & (90.9, 74.6) & (88.1, 89.7) & (\textbf{96.4}, 89.3) \\
        
        \rowcolor{blue!25}
        \cmark & \cmark & \cmark & \cmark & \cmark & (92.4, 96.6) & \textbf{(91.9, 85.7)} & \textbf{(89.5, 90.7)} & (96.3, \textbf{91.4}) \\
    \bottomrule
    \end{tabular}
    }
    \label{tab:ablation}
\end{table}

% \begin{figure*}[htbp]
%   \centering
%   \includegraphics[width=\textwidth]{pic/pic_sub_and_view.pdf}
%   \caption{Overall visualization of four image sets demonstrating the effects of the proposed modules. (a) Multi-View Foreground-Background Enhancement (MVFBE) results, highlighting the enhancement of tokens across multiple views. (b) Background Subtraction (BS) process, showing how irrelevant background tokens are removed to emphasize anomalies. Together, these visualizations illustrate the progressive improvement in anomaly representation and localization.}
%   \label{fig:mutil-view-Back-sub}
% \end{figure*}

\subsubsection{Effectiveness of MSTFF}

The inclusion of our MSTFF module (column \textbf{MSTFF} in Table~\ref{tab:ablation}) leads to substantial improvements over the \textbf{Base} model. On MVTec, image-level AUROC increases from 67.1\% to 82.0\%, and pixel-level PRO rises from 32.9\% to 68.0\%, demonstrating that MSTFF effectively captures richer semantic cues and task-relevant information for more accurate anomaly recognition and localization. Additional modules—\textbf{MVFBE}, \textbf{BS}, and \textbf{SCR}—provide further gains, and the full configuration combining all four strategies achieves the best overall performance on both MVTec and VisA. These results confirm that MSTFF plays a \textbf{critical role} while synergizing with the other components. Additional ablation experiments can be found in Appendix~\ref{ab_MSTFF}.

\subsubsection{Effectiveness of MVFBE}

The addition of the MVFBE module (column \textbf{MVFBE} in Table~\ref{tab:ablation}) further improves performance over the \textbf{Base} and \textbf{MSTFF} configurations. On MVTec, it increases image-level AUROC to 92.2\% and pixel-level PRO to 84.5\%, indicating that MVFBE effectively enhances the distinction between foreground anomalies and background context. Similar gains are observed on VisA, showing that MVFBE consistently boosts anomaly recognition and localization by emphasizing relevant object regions across multiple views.

\begin{figure*}[htbp]
  \centering
  \includegraphics[width=\textwidth]{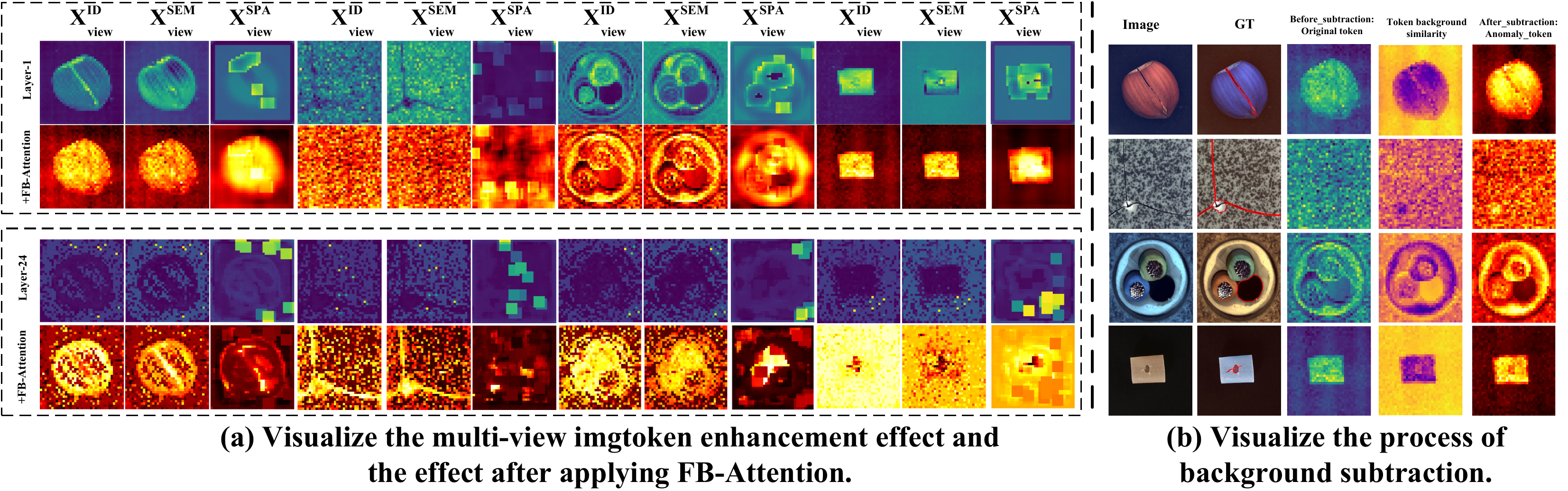}
  \caption{Overall visualization of four image sets demonstrating the effects of the proposed modules. (a) Multi-View Foreground-Background Enhancement (MVFBE) results, highlighting the enhancement of tokens across multiple views. (b) Background Subtraction (BS) process, showing how irrelevant background tokens are removed to emphasize anomalies. Together, these visualizations illustrate the progressive improvement in anomaly representation and localization.}
  \label{fig:mutil-view-Back-sub}
\end{figure*}

In Figure~\ref{fig:mutil-view-Back-sub}(a), we visualize the effect of our multi-view imgtoken enhancement and the subsequent FB-Attention. The top row shows the first layer imgtokens obtained from the CLIP pre-trained ViT model, while the middle row displays the features from the 24th layer: $\mathbf{X}^{\text{ID}}$ along with the enhanced $\mathbf{X}^{\text{SEM}}$ and $\mathbf{X}^{\text{SPA}}$ views. Comparing the results before and after applying FB-Attention, it is evident that FB-Attention effectively suppresses background noise and enhances the foreground anomaly regions. This improvement is especially pronounced for complex textures and small defects, where the enhanced attention maps provide clearer localization and stronger contrast between anomalous objects and the background, demonstrating the benefit of integrating multi-view enhancement with FB-Attention. The block-like artifacts in the SPA view arise from noise smoothing and signal enhancement, forming connected regions, which aligns with the assumption that anomalies are connected and does not affect the capture of fine-grained local structures. Additional ablation experiments  in Appendix~\ref{ab_MVFBE}.

\subsubsection{Effectiveness of Background Suppression}

To assess the impact of token-level background subtraction, we visualize the intermediate token representations during the anomaly localization process in Figure~\ref{fig:mutil-view-Back-sub}(b). The first column shows the original image, the second column presents the ground truth, and the subsequent columns illustrate the background subtraction pipeline. Specifically, the “Before Subtraction” column displays the original token representations, the “Token Background Similarity” maps indicate the similarity between each token and the background tokens, and the “After Subtraction” column highlights the anomaly tokens after background removal. Darker or brighter regions in the similarity maps correspond to higher background resemblance, providing an interpretable cue for the subtraction process.

We observe that the original tokens often mix target and background information, which can obscure subtle anomalies. By leveraging background similarity, irrelevant regions are effectively suppressed, and the resulting anomaly tokens clearly emphasize defects or critical structures. Small scratches become distinctly visible, anomalies embedded in complex patterns are accurately separated, and minute anomaly points are highlighted, while noise is substantially reduced. These results demonstrate that token-level background subtraction enables the model to focus on salient regions without discarding essential features, thereby improving the clarity and reliability of anomaly localization.

\subsubsection{Effectiveness of SCR}

To evaluate the effectiveness of our SCR module (column \textbf{SCR} in Table~\ref{tab:ablation}), we conduct a detailed ablation study on both MVTec and VisA datasets. When applied individually, SCR shows limited improvement on image-level anomaly detection for MVTec (AUROC/AP: 65.4/81.3 vs.\ 67.1/84.0 for the Base), but substantially enhances pixel-level anomaly localization, with PRO increasing from 32.9 to 45.5. On VisA, SCR improves both image-level (AUROC/AP: 63.8/69.0 vs.\ 45.7/56.2) and pixel-level metrics, demonstrating its effectiveness in refining fine-grained anomaly regions.
% When combined with other modules, SCR exhibits strong complementary effects.
In the full configuration integrating all four strategies, it contributes to the highest image-level AUROC/AP (89.5/90.7) on VisA, outperforming all other combinations. These results indicate that SCR synergizes with the other modules to improve both global anomaly recognition and local anomaly localization, validating its role in enforcing spatial consistency and boosting fine-grained anomaly detection.

% \textbf{Additional details regarding the ablation experiments can be found in the supplementary material.}

%% file: sec/5_condlusion.tex
\section{Conclusion}

In this work, we propose \textbf{FB-CLIP}, a unified framework for zero-shot anomaly detection that achieves fine-grained and robust anomaly representations through collaborative improvements in both textual and visual modalities. In the text modality, FB-CLIP fuses EOT, global-pooled, and attention-weighted token features to construct rich, task-aware text representations. It further introduces \textit{Semantic Consistency Regularization}, which minimizes entropy and enforces similarity margin constraints to suppress uncertain matches and enhance cross-modal alignment stability. In the visual modality, FB-CLIP leverages multi-view foreground-background soft separation and continuous gating to integrate multi-branch attention outputs, emphasizing critical anomalous cues, while \textit{Background Suppression} reduces complex scene and cross-domain interference to produce cleaner and more robust visual representations. By integrating these strategies, FB-CLIP effectively captures fine-grained local anomaly patterns while maintaining robust ZSAD across diverse datasets and domains.

%% file: sec/X_suppl.tex
% \clearpage
% \setcounter{page}{1}
% \maketitlesupplementary

\clearpage
\setcounter{page}{1}
\renewcommand{\thesection}{\Alph{section}}  % 将章节编号改为 A, B, C ...
\onecolumn                                   % 切换为单栏模式
\setcounter{section}{0}                      % 重置章节计数
\maketitlesupplementary

\begin{figure}[htbp]
\centering
\includegraphics[width=\linewidth]{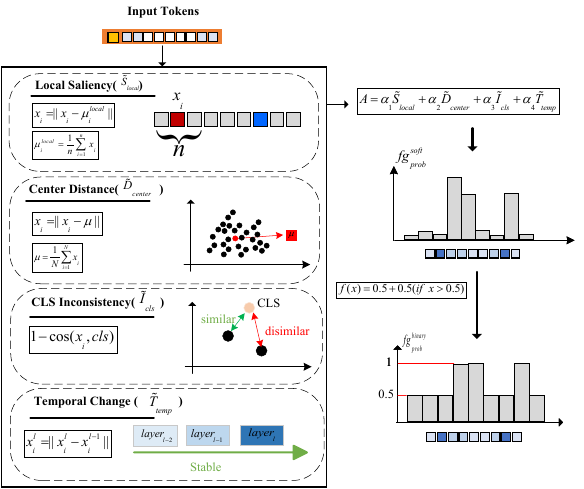}
\caption{Generation process of the foreground–background mask. Multiple complementary anomaly indicators are first aggregated to estimate the foreground–background probability of each token.}
\label{fig:mvfbe_mask}
\end{figure}

\section{Details of Multi-View Foreground-Background Enhancement}
\label{soft_mask}

This section provides detailed formulations of the indicators used in the Multi-View Foreground-Background Enhancement (MVFBE) module described in Sec.~X.

Given visual tokens $\mathbf{X} \in \mathbb{R}^{B \times (L+1) \times C}$ extracted from a Vision Transformer, including a class token $\mathbf{x}_{\text{cls}}$ and patch tokens $\mathbf{x}_{\text{img}}$, MVFBE estimates foreground likelihood using four complementary indicators.

\textbf{Local saliency.} Foreground regions typically exhibit stronger feature deviation from their local context compared with homogeneous background regions. Inspired by prior studies on token-level saliency estimation in Vision Transformers~\cite{xie2022vit, dong2023heatvit, liu2021visual}, we quantify the local distinctiveness of each patch token by measuring its deviation from a neighborhood-aggregated representation. Specifically, we construct a local contextual representation using 1D average pooling with kernel size 3 along the token sequence. The local saliency score is defined as the $\ell_2$ distance between each token and its neighborhood average: \begin{equation} \mathcal{S}{\text{local}} = \left| \mathbf{x}{\text{img}} - \text{AvgPool}(\mathbf{x}_{\text{img}}) \right|_2 . \end{equation} This formulation highlights tokens whose representations significantly differ from their surrounding context, which often correspond to potential foreground structures.

\textbf{Center distance.}
Foreground tokens tend to deviate from the dominant feature distribution formed by background regions. Following the common practice of distribution-based anomaly or foreground modeling~\cite{ruff2018deep}, we characterize this deviation by computing the distance between each token and a global distribution center.

Formally, the center distance indicator is defined as

\begin{equation}
\mathcal{D}{\text{center}} =
\left|
\mathbf{x}{\text{img}} - \mathbf{c}
\right|_2 ,
\end{equation}

where the center $\mathbf{c}$ can either be a predefined normal prototype $\mathbf{c}_{\text{normal}}$ or the empirical batch mean:

\begin{equation}
\mathbf{c} =
\frac{1}{L}\sum_{i=1}^{L} \mathbf{x}_{\text{img}}^{(i)} .
\end{equation}

Tokens that lie far from the global center are more likely to correspond to foreground or abnormal structures.

\textbf{CLS Inconsistency.} 
In Vision Transformers, the class token (CLS token) aggregates global semantic information across the entire image through self-attention~\cite{chefer2021transformer, wu2020visual}. Therefore, patch tokens that are semantically inconsistent with the CLS token are likely to correspond to localized foreground or anomalous regions.

To quantify this discrepancy, we compute the cosine dissimilarity between each patch token and the CLS token:

\begin{equation}
\mathcal{I}_{\text{cls}}^{(i)}
=
1 - \text{cosine\_similarity}
\big(\mathbf{x}_{\text{img}}^{(i)}, \mathbf{x}_{\text{cls}}\big),
\end{equation}

where $\mathbf{x}_{\text{img}}^{(i)}$ denotes the $i$-th patch token and $\mathbf{x}_{\text{cls}}$ denotes the global class token. A higher inconsistency score indicates a greater semantic deviation from the global image representation, highlighting potential foreground structures.

\textbf{Temporal variation.}
Deep Transformer layers progressively refine token representations through hierarchical feature transformations~\cite{peng2024d, dosovitskiy2020image}. Tokens corresponding to foreground structures often exhibit larger representation shifts across layers due to progressive semantic enrichment.

To capture this dynamic behavior, we measure the feature variation between adjacent layers:

\begin{equation}
\mathcal{T}_{\text{temp}} =
|
\mathbf{X}^{(l)} - \mathbf{X}^{(l-1)}
|_2 .
\end{equation}

This temporal variation reflects the degree of representation evolution during hierarchical feature refinement.

\textbf{Indicator normalization.}
To ensure fair contribution from each cue, we apply per-sample min–max normalization:

\begin{equation}
\tilde{\mathcal{S}}_i =
\frac{\mathcal{S}_i - \min_j(\mathcal{S}_i[j])}
{\max_j(\mathcal{S}_i[j]) - \min_j(\mathcal{S}_i[j]) + \epsilon}.
\end{equation}

% All normalized indicators are then aggregated as described in Eq.~(X) in the main paper.
The normalized indicators are then linearly combined to form an anomaly score:
\begin{equation}
 \mathcal{A} = \alpha_1 \tilde{\mathcal{S}}_{\text{local}} + \alpha_2 \tilde{\mathcal{D}}_{\text{center}} + \alpha_3 \tilde{\mathcal{I}}_{\text{cls}} + \alpha_4 \tilde{\mathcal{T}}_{\text{temp}},
\end{equation}

% \clearpage

\input{sec/FG_BG}

\begin{figure*}[!b]
    \centering
    \includegraphics[width=\textwidth]{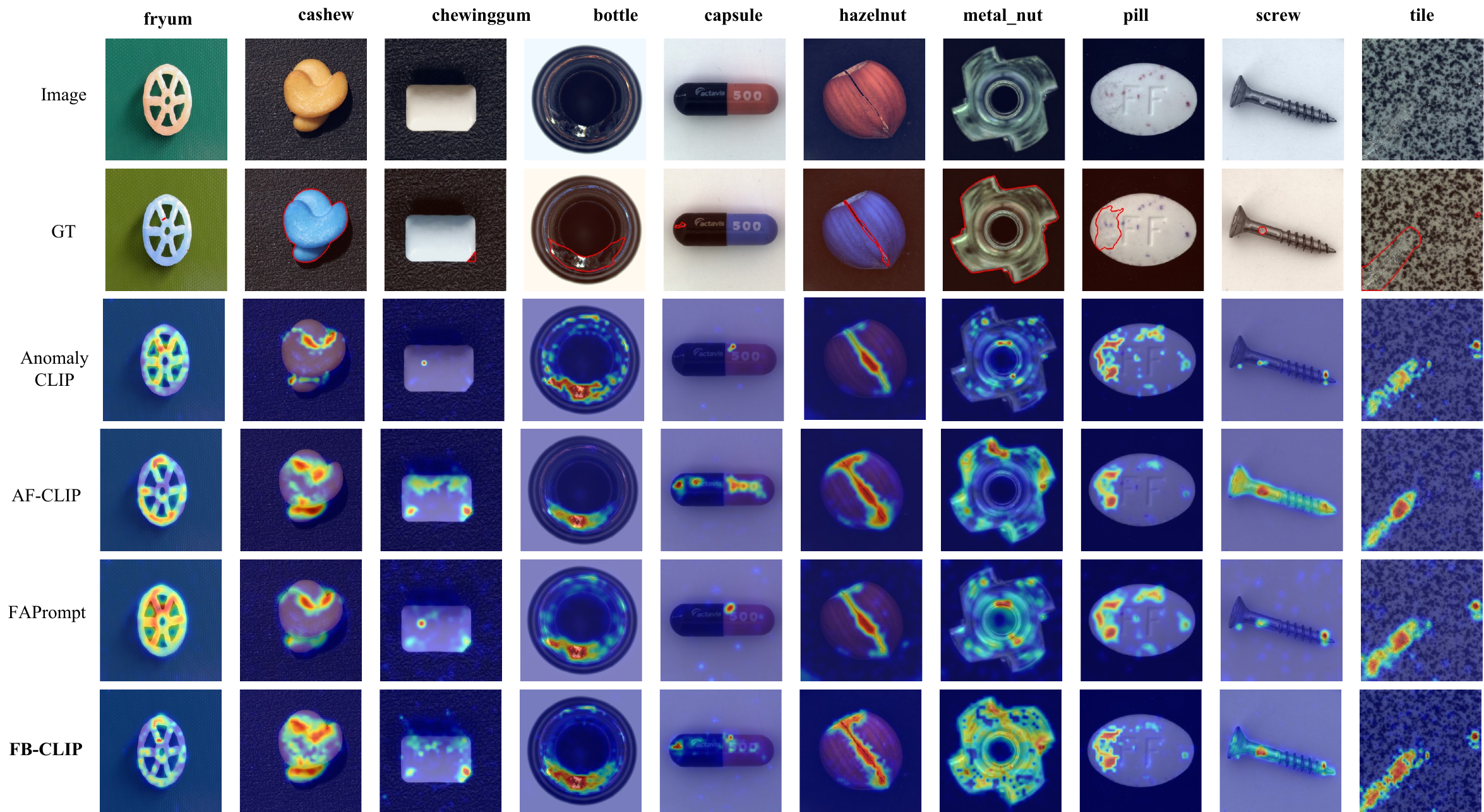} % 修改为你的图片路径
    \caption{
    Visualization results of anomaly localization across multiple categories using different CLIP-based methods. 
    }
    \label{fig:fbclip_comparison}
\end{figure*}

\section{Visual comparison with other state-of-the-art methods}
\label{view_sota}

Figure~\ref{fig:fbclip_comparison} presents a qualitative comparison of anomaly localization results 
on various product categories, including \textit{fryum}, \textit{cashew}, 
\textit{chewinggum}, \textit{bottle}, \textit{capsule}, \textit{hazelnut}, \textit{metal\_nut}, 
\textit{pill}, \textit{screw}, and \textit{tile}. 
The first row displays the original images, and the second row provides the ground-truth (GT) anomaly regions. 
Subsequent rows correspond to the results obtained by Anomaly CLIP, AF-CLIP, FAPrompt, and our proposed \textbf{FB-CLIP}.

As shown in the figure, \textbf{FB-CLIP} achieves more precise and concentrated anomaly localization results 
that align closely with the ground-truth annotations. 
Compared with other methods, it produces fewer false activations in background regions and captures subtle defects more effectively. 
For instance, in the \textit{hazelnut} and \textit{metal\_nut} samples, FB-CLIP successfully highlights the entire defective region, 
while the other methods show incomplete or scattered activations. 
On fine-grained textures such as \textit{tile}, FB-CLIP also demonstrates superior robustness and sensitivity to small-scale anomalies, 
indicating its stronger generalization capability across diverse categories.

\section{Ablation Study of Multi-Strategy Text Feature Fusion (MSTFF)}
\label{ab_MSTFF}

\begin{table*}[htbp]
\centering
\caption{
Ablation study on \textbf{Multi-Strategy Text Feature Fusion (MSTFF)} 
across two datasets (\textbf{MVTec} and \textbf{VisA}). 
EOT: End-of-Text; GP: Global Pooling; Attn: Attention-weighted.
}
\label{tab:mstff_ablation}
\setlength{\tabcolsep}{5pt}
\renewcommand{\arraystretch}{1.2}
\resizebox{\textwidth}{!}{
\begin{tabular}{ccccccc}
\toprule
\multicolumn{3}{c}{\textbf{Strategies}} 
& \multicolumn{2}{c}{\textbf{MVTec}} 
& \multicolumn{2}{c}{\textbf{VisA}} \\

\cmidrule(lr){1-3}
\cmidrule(lr){4-5}
\cmidrule(lr){6-7}

\textbf{EOT} & \textbf{GP} & \textbf{Attn}
& \shortstack{\textbf{Image-level}\\(AUROC, AP)}
& \shortstack{\textbf{Pixel-level}\\(AUROC, PRO)}
& \shortstack{\textbf{Image-level}\\(AUROC, AP)}
& \shortstack{\textbf{Pixel-level}\\(AUROC, PRO)} \\

\midrule

\checkmark &  &  & (67.1, 84.0) & (65.2, 32.9) & (45.7, 56.2) & (88.4, 71.8) \\

 & \checkmark &  & (62.6, 81.9) & (63.3, 31.9) & (51.3, 56.6) & (83.7, 67.5) \\

 &  & \checkmark & (80.9, 90.8) & (79.0, 55.7) & (78.5, \textbf{80.7}) & (92.7, 80.3) \\

\checkmark &  & \checkmark & (81.1, 90.9) & (83.1, 66.4) & \textbf{(78.6, 80.7)} & (92.7, 80.4) \\

\checkmark & \checkmark & \checkmark 
& \textbf{(82.0, 91.4)} & \textbf{(84.0, 68.0)} 
& (78.3, 80.5) & \textbf{(93.2, 81.4)} \\

\bottomrule
\end{tabular}
}
\end{table*}

Table~\ref{tab:mstff_ablation} presents an ablation study of our proposed \textbf{Multi-Strategy Text Feature Fusion (MSTFF)} on two datasets (\textbf{MVTec} and \textbf{VisA}), evaluated at both image-level (AUROC and AP) and pixel-level (AUROC and PRO). The study compares single-strategy configurations (EOT, Global, Attention) against multi-strategy combinations (EOT+Attn, EOT+GP+Attn), allowing for a detailed assessment of the contribution of each textual feature and their combinations to anomaly detection and localization performance.

Considering single-strategy results, using only \textbf{EOT} or \textbf{Global} features yields limited performance. On MVTec, EOT achieves an image-level AUROC of 67.1 and AP of 84.0, but its pixel-level AUROC and PRO are only 65.2 and 32.9, respectively. Global features show similar trends, with an image-level AUROC of 62.6 and modest improvements in pixel-level metrics. These results indicate that relying solely on text-end features or global pooling is insufficient to capture fine-grained anomalies. In contrast, the \textbf{Attention} feature significantly improves performance across both datasets. On MVTec, it achieves an image-level AUROC of 80.9 and pixel-level AUROC of 79.0; on VisA, the image-level AUROC reaches 78.5 and pixel-level AUROC 92.7. This demonstrates that attention-weighted features effectively focus on the most relevant regions, enhancing localization precision.

Multi-strategy combinations further boost performance. Integrating \textbf{EOT with Attention (EOT+Attn)} increases the MVTec pixel-level PRO from 55.7 to 66.4 compared to Attention alone, indicating that EOT provides complementary information that enhances sensitivity to fine-grained anomalies. Incorporating global pooling as well (\textbf{EOT+GP+Attn}) yields the best overall performance: image-level AUROC reaches 82.0 and pixel-level AUROC 84.0 on MVTec, while VisA pixel-level PRO increases to 81.4. These findings suggest that fusing multiple textual feature strategies allows MSTFF to leverage complementary information effectively, resulting in more precise and robust anomaly detection and localization.

In summary, the ablation results clearly validate the effectiveness of MSTFF. While single-strategy configurations provide limited anomaly cues, attention-weighted features and multi-strategy fusion substantially improve both image-level and pixel-level metrics. The notable gains in pixel-level PRO highlight MSTFF’s ability to capture fine-grained defect regions and its superior generalization across diverse datasets.

\begin{table*}[htbp]
\centering
\caption{
Ablation study with modular configuration using four components:
 \textbf{ID}, \textbf{SEM},\textbf{SPA} and \textbf{FB-Attention},.
A checkmark (\checkmark) indicates that the component is enabled.
}
\label{tab:module_ablation}
\setlength{\tabcolsep}{12pt}
\renewcommand{\arraystretch}{1.25}
\resizebox{\textwidth}{!}{
\begin{tabular}{cccccccc}
\toprule

\multicolumn{4}{c}{\textbf{Method}} &
\multicolumn{2}{c}{\textbf{MVTec}} &
\multicolumn{2}{c}{\textbf{VisA}} \\
\cmidrule(lr){1-4} \cmidrule(lr){5-6} \cmidrule(lr){7-8}
\textbf{ID} &
\textbf{SEM} &
\textbf{SPA} & 
\textbf{FB-Attn} &
\textbf{Img} (AUROC, AP) &
\textbf{Pix} (AUROC, PRO) &
\textbf{Img} (AUROC, AP) &
\textbf{Pix} (AUROC, PRO) \\
\midrule

 \checkmark & &   &   & (82.0, 91.4) & (84.0, 68.0) & (78.3, 80.5) & (93.2, 81.4) \\
& \checkmark &   &   & (83.0, 92.2) & (77.3, 48.6) & (80.8, 82.9) & (92.1, 77.1) \\
& &  \checkmark   &   & (83.0, 91.9) & (80.8, 55.6) & (79.3, 81.7) & (91.0, 73.9) \\ 

 \checkmark &  \checkmark &   &   & (81.4, 91.1) & (84.8, 67.1) & (79.1, 81.5) & (93.4, 81.7) \\

 \checkmark &   & \checkmark  &   & (82.0, 91.3) & (83.1, 64.3) & (77.8, 80.5) & (93.3, 80.9) \\

 &  \checkmark   & \checkmark  &   & (81.6, 91.2) & (81.6, 61.2) & (78.8, 81.3) & (92.9, 79.3) \\

\checkmark  &  \checkmark   & \checkmark  &   & (81.7, 91.4) & (83.7, 64.8) & (78.8, 81.0) & (93.3, 81.1) \\
% FB-Attention

\checkmark &   &   & \checkmark  & (92.5, 96.7) & (90.8, 47.9) & (86.7, 88.6) & (95.8, 87 .8) \\

  & \checkmark  &   & \checkmark  & \textbf{(92.7, 96.7)} & (86.2, 55.1) & (85.6, 87.5) & (95.7, 87.4) \\

  &  & \checkmark   & \checkmark  & (92.2, 96.3) & (90.2, 81.8) & (88.2, 89.4) & (94.4, 83.8) \\
    \checkmark & \checkmark  &    & \checkmark  & (91.7, 96.1) & (91.4, 85.1) & (87.2, 88.8) & (95.9, 90.1) \\

 \checkmark &  & \checkmark   & \checkmark  & (92.3, 96.5) & (90.9, 83.8) & (88.2, 89.6) & (\textbf{96.0}, 90.8) \\

   & \checkmark  &   \checkmark   & \checkmark  & (92.0, 96.3) & (\textbf{91.7}, 82.8) & (88.1, 89.4) & (95.6, 87.8) \\

  \checkmark  & \checkmark  &   \checkmark   & \checkmark  & (92.2, 96.5) & \textbf{(91.7, 84.5)} & \textbf{(88.4, 89.7)} & \textbf{(96.0, 90.8)} \\ 

\bottomrule
\end{tabular}
}
\end{table*}

\begin{table*}[htbp]
\centering
\small
\setlength{\tabcolsep}{23pt}
\renewcommand{\arraystretch}{1.1}

\caption{Ablation study on the number of tokens for background representation. 
Performance is reported on MVTec and VisA datasets, including image-level (I) AUROC and AP, 
as well as pixel-level (P) AUROC and PRO metrics.}

\label{tab:token_ablation}

\begin{tabular}{lcccc}
\toprule
 & \multicolumn{2}{c}{\textbf{MVTec}} 
 & \multicolumn{2}{c}{\textbf{VisA}} \\
\cmidrule(lr){2-3} \cmidrule(lr){4-5}
 & I-(AUROC, AP) 
 & P-(AUROC, PRO) 
 & I-(AUROC, AP) 
 & P-(AUROC, PRO) \\
\midrule
L   & (92.4, 96.6) & (91.8, 85.4) & (89.3, 90.5) & (96.3, 91.4) \\
L/2 & (92.4, 96.6) & (91.9, 85.7) & (89.5, 90.7) & (96.3, 91.4) \\
\bottomrule
\end{tabular}

\end{table*}

\section{Ablation Study on MVFBE Components}
\label{ab_MVFBE}

To validate the effectiveness of each component within the proposed \textbf{Multi-View Foreground-Background Enhancement (MVFBE)} module (Figure~\ref{fig:FBE}), we conduct a modular ablation study, the results of which are reported in Table~\ref{tab:module_ablation}. The MVFBE module consists of four complementary components: \textbf{Identity (ID)}, \textbf{Semantic (SEM)}, \textbf{Spatial (SPA)}, and \textbf{Foreground-Background Attention (FB-Attn)}. Each component can be independently enabled to examine its contribution to both image-level (AUROC, AP) and pixel-level (AUROC, PRO) anomaly detection performance across two datasets, \textbf{MVTec} and \textbf{VisA}.  

From the single-component configurations, it is evident that \textbf{FB-Attn} alone significantly boosts image-level performance, achieving AUROC/AP of (92.5, 96.7) on MVTec and (86.7, 88.6) on VisA. However, its pixel-level PRO on MVTec is relatively low (47.9), suggesting that while FB-Attn effectively enhances global feature discrimination, precise pixel-level localization benefits from complementary components. In contrast, the traditional enhancement views (ID, SEM, SPA) individually provide moderate improvements. For instance, ID alone yields (82.0, 91.4) image-level AUROC/AP and (84.0, 68.0) pixel-level AUROC/PRO on MVTec, indicating that preserving original features maintains baseline information but lacks strong localization capability. SEM emphasizes semantic information, improving image-level AUROC (83.0) but achieving lower pixel-level PRO (48.6), highlighting its role in capturing foreground richness while background stability remains less effectively separated. SPA focuses on fine-grained spatial structures, achieving balanced improvements in both image- and pixel-level metrics, confirming its utility in capturing local contextual anomalies.

When combining traditional enhancement views, synergistic effects emerge. The combination of ID+SEM+SPA without FB-Attn increases pixel-level PRO to 64.8 on MVTec and 81.1 on VisA, while maintaining strong image-level AUROC/AP. This demonstrates that complementary perspectives contribute to more accurate anomaly localization by aggregating multiple views of foreground and background features.  

Integrating \textbf{FB-Attn} with one or more enhancement views leads to substantial performance gains across all metrics. Notably, the full combination of ID+SEM+SPA+FB-Attn achieves the highest overall performance, with MVTec image-level AUROC/AP of (92.2, 96.5) and pixel-level AUROC/PRO of (91.7, 84.5), and VisA image-level (88.4, 89.7) and pixel-level (96.0, 90.8). This confirms that FB-Attn effectively refines the aggregated features from multiple views, enhancing both global detection and fine-grained localization. The consistent improvements in pixel-level PRO when FB-Attn is included emphasize its critical role in discriminating foreground anomalies from background regions, complementing the ID, SEM, and SPA components.  

In summary, the ablation results validate that each component of MVFBE contributes distinct and complementary strengths: ID preserves raw feature integrity, SEM models foreground richness and background stability, SPA captures local spatial structures, and FB-Attn refines these features through attention-based foreground-background separation. The full combination of all four components yields the most balanced and robust performance, demonstrating the effectiveness of the proposed modular design in both image-level recognition and pixel-level anomaly localization.

\section{Impact of Token Number on Background Representation}

We investigate how the number of tokens affects background representation by comparing the performance using $L$ tokens versus $L/2$ tokens. Table~\ref{tab:mstff_ablation} reports the results on MVTec and VisA datasets, including image-level (I) AUROC and AP, as well as pixel-level (P) AUROC and PRO metrics.  

As shown, using $L/2$ tokens achieves virtually identical performance to using $L$ tokens. On MVTec, the image-level AUROC and AP remain the same at 92.4\% and 96.6\%, while the pixel-level AUROC slightly improves from 91.8\% to 91.9\% and PRO increases from 85.4 to 85.7. On VisA, $L/2$ tokens maintain comparable or slightly better performance for image-level (AUROC/AP: 89.5/90.7 vs.\ 89.3/90.5) and identical pixel-level metrics.  

These results demonstrate that half the number of tokens is sufficient to effectively capture background information. Reducing token count not only maintains or slightly improves anomaly detection and localization performance but also provides potential computational savings, making it an efficient strategy for multi-strategy text feature fusion and multi-view foreground-background enhancement.

% \begin{table}[htbp]
% \centering
% \scriptsize
% \setlength{\tabcolsep}{1pt}
% \renewcommand{\arraystretch}{0.9}
% \caption{Ablation study on the number of tokens for background representation. Performance is reported on MVTec and VisA datasets, including image-level (I) AUROC and AP, as well as pixel-level (P) AUROC and PRO metrics.}
% \label{tab:mstff_ablation}
% \begin{tabular}{lcccc}
% \toprule
%  & \multicolumn{2}{c}{\textbf{MVTec}} 
%  & \multicolumn{2}{c}{\textbf{VisA}} \\
% \cmidrule(lr){2-3} \cmidrule(lr){4-5}
%  & I-(AUROC, AP) 
%  & P-(AUROC, PRO) 
%  & I-(AUROC, AP) 
%  & P-(AUROC, PRO) \\
% \midrule
% L  & (92.4, 96.6) & (91.8, 85.4) & (89.3, 90.5) & (96.3, 91.4) \\
% L/2  & (92.4, 96.6) & (91.9, 85.7) & (89.5, 90.7) & (96.3, 91.4) \\
% \bottomrule
% \end{tabular}
% \end{table}

% \begin{table*}[t]
% \centering
% \small
% \setlength{\tabcolsep}{6pt}
% \renewcommand{\arraystretch}{1.1}

% \caption{Ablation study on the number of tokens for background representation. 
% Performance is reported on MVTec and VisA datasets, including image-level (I) AUROC and AP, 
% as well as pixel-level (P) AUROC and PRO metrics.}

% \label{tab:token_ablation}

% \begin{tabular}{lcccc}
% \toprule
%  & \multicolumn{2}{c}{\textbf{MVTec}} 
%  & \multicolumn{2}{c}{\textbf{VisA}} \\
% \cmidrule(lr){2-3} \cmidrule(lr){4-5}
%  & I-(AUROC, AP) 
%  & P-(AUROC, PRO) 
%  & I-(AUROC, AP) 
%  & P-(AUROC, PRO) \\
% \midrule
% L   & (92.4, 96.6) & (91.8, 85.4) & (89.3, 90.5) & (96.3, 91.4) \\
% L/2 & (92.4, 96.6) & (91.9, 85.7) & (89.5, 90.7) & (96.3, 91.4) \\
% \bottomrule
% \end{tabular}

% \end{table*}

\clearpage
\section{Subject-level}
\label{sec:Subject-level_results}
\input{sec/subject_result}

\clearpage
\section{Limitations}

\input{sec/LimitationLimitation}

\clearpage
\section{Anomaly Detection in Challenging Scenarios with Physical Occlusion}

\input{sec/Occusion}

%% file: sec/FG_BG.tex
\section{FG/BG Token Enhancement Methods}
\label{MVFBE_enhance}

This section presents two token enhancement strategies for foreground/background (FG/BG) refinement: \textbf{Semantic Enhancement (SEM)} and \textbf{Spatial Enhancement (SPA)}. Both methods aim to refine token features based on foreground probabilities, but they differ in focus: SEM emphasizes global semantic relations, while SPA emphasizes local spatial consistency.

\subsection{Semantic Enhancement (SEM)}
\label{code_sem}

\textbf{Core Idea}: Use the CLS token to capture global semantic information and reweight tokens based on their foreground probability and semantic richness. Foreground tokens receive higher weights if they are semantically informative (less similar to CLS), while background tokens are weighted for stability.

\begin{algorithm} 
\SetAlgoLined
\KwIn{tokens $X \in \mathbb{R}^{b \times (l+1) \times c}$, foreground probabilities $P_{fg} \in \mathbb{R}^{b \times l}$}
\KwOut{Refined tokens $X_{out}$}

\textcolor{blue}{\textit{Extract CLS token:}} $CLS = X[:,0,:]$\;
\textcolor{blue}{\textit{Extract other tokens:}} $X_{tokens} = X[:,1:,:]$\;
\textcolor{blue}{\textit{Compute foreground weight matrix:}} $W_{fg} = P_{fg} \cdot P_{fg}^T$\;
\textcolor{blue}{\textit{Compute background weight matrix:}} $W_{bg} = (1 - P_{fg}) \cdot (1 - P_{fg})^T$\;
\textcolor{blue}{\textit{Compute similarity to CLS:}} $S = \text{cosine\_similarity}(X_{tokens}, CLS)$\;
\textcolor{blue}{\textit{Compute information richness:}} $R = 1 - S$\;
\textcolor{blue}{\textit{Compute foreground aggregation weights:}} $A_{fg} = \text{softmax}(R \cdot W_{fg})$\;
\textcolor{blue}{\textit{Compute background aggregation weights:}} $A_{bg} = \text{softmax}(S \cdot W_{bg})$\;
\textcolor{blue}{\textit{Aggregate tokens:}} $X_{fg\_agg} = A_{fg} \cdot X_{tokens}, \ X_{bg\_agg} = A_{bg} \cdot X_{tokens}$\;
\textcolor{blue}{\textit{Combine foreground and background:}} $X_{agg} = \alpha (X_{fg\_agg} + X_{bg\_agg}) + (1-\alpha) X_{tokens}$\;
\textcolor{blue}{\textit{Concatenate CLS token:}} $X_{out} = [CLS, X_{agg}]$\;

\caption{SEM Enhancement Pseudocode}
\end{algorithm}

\textbf{Analysis}:
\begin{itemize}
    \item Captures global semantic relations between tokens.
    \item Highlights tokens that are semantically informative (foreground).
    \item Background tokens are stabilized using similarity to CLS.
    \item Computational complexity scales with $O(l^2)$ due to pairwise token weighting.
\end{itemize}

\subsection{Spatial Enhancement (SPA)}
\label{code_spa}

\textbf{Core Idea}: Incorporate spatial structure by reshaping tokens into 2D patches and aggregating local neighborhoods. Foreground and background are weighted based on local stability and information richness.

\begin{algorithm}
\caption{SPA Enhancement Pseudocode}
\KwIn{tokens $X \in \mathbb{R}^{b \times (l+1) \times c}$, foreground probabilities $P_{fg} \in \mathbb{R}^{b \times l}$, patch size $r$}
\KwOut{Refined tokens $X_{out}$}

\textcolor{blue}{\textit{Extract CLS token:}} $CLS = X[:,0,:]$\;
\textcolor{blue}{\textit{Extract other tokens:}} $X_{tokens} = X[:,1:,:]$\;
\textcolor{blue}{\textit{Reshape tokens to 2D:}} $X_{2D} = \text{reshape}(X_{tokens}, [b, c, h, w]), \ h = w = \sqrt{l}$\;
\textcolor{blue}{\textit{Unfold patches:}} $X_{patch} = \text{unfold}(X_{2D}, \text{kernel}=r)$\;
\textcolor{blue}{\textit{Unfold foreground mask:}} $P_{fg\_patch} = \text{unfold}(P_{fg})$\;
\textcolor{blue}{\textit{Compute background mask:}} $P_{bg\_patch} = 1 - P_{fg\_patch}$\;
\textcolor{blue}{\textit{Compute stability score for background:}} $S_{bg} = \text{stability}(X_{patch} \cdot P_{bg\_patch}, CLS)$\;
\textcolor{blue}{\textit{Compute information richness for foreground:}} $R_{fg} = \text{richness}(X_{patch} \cdot P_{fg\_patch}, CLS)$\;
\textcolor{blue}{\textit{Compute weighted aggregation:}} $X_{bg\_agg} = \sum (X_{patch} \cdot \text{softmax}(S_{bg})), \ X_{fg\_agg} = \sum (X_{patch} \cdot \text{softmax}(R_{fg}))$\;
\textcolor{blue}{\textit{Combine foreground and background:}} $X_{agg} = X_{fg\_agg} + X_{bg\_agg}$\;
\textcolor{blue}{\textit{Reshape to original token layout and concatenate CLS token:}} $X_{out} = [CLS, X_{agg}]$\;

\end{algorithm}

\textbf{Analysis}:
\begin{itemize}
    \item Preserves spatial consistency by aggregating local patches.
    \item Foreground and background are weighted based on local patch statistics.
    \item Sensitive to token layout (requires $l$ to be square for 2D reshaping).
    \item Computational complexity scales with $O(b \cdot h \cdot w \cdot r^2)$, depending on patch size.
\end{itemize}

\textbf{Summary}:
\begin{itemize}
    \item SEM emphasizes \textit{global semantic relations}, effective for highlighting informative tokens across the image.
    \item SPA emphasizes \textit{local spatial consistency}, effective for preserving structure and local context.
    \item Both methods can be complementary in a FG/BG token refinement framework.
\end{itemize}

%% file: sec/subject_result.tex
To provide a more detailed evaluation, we report the subset–level performance in the following tables (Table~\ref{tab:dagm_results}, \ref{tab:mpdd_results}, \ref{tab:btad_results}, \ref{tab:DTD}, \ref{tab:visa_to_mvtec}, \ref{tab:mvtec_to_visa}, \ref{tab:realIAD_results}
).

\begin{table}[htbp]
    \centering
    \caption{Performance on the DAGM\_KAGGLEUPLOAD dataset.}
    \setlength{\tabcolsep}{4pt}
    \renewcommand{\arraystretch}{1.15}
    \resizebox{\columnwidth}{!}{
    \begin{tabular}{lcccc}
    \toprule
        \multirow{2}{*}{\textbf{Objects}} 
        & \multicolumn{2}{c}{\textbf{Pixel}} 
        & \multicolumn{2}{c}{\textbf{Image}} \\
        \cmidrule(lr){2-3} \cmidrule(lr){4-5}
        & \textbf{AUROC} & \textbf{AUPRO} & \textbf{AUROC} & \textbf{AP} \\
    \midrule
        Class1  & 91.8 & 82.1 & 94.1 & 82.1 \\
        Class2  & 99.7 & 99.0 & 99.9 & 99.8 \\
        Class3  & 97.0 & 96.3 & 100.0 & 100.0 \\
        Class4  & 94.0 & 80.5 & 99.5 & 97.1 \\
        Class5  & 99.3 & 97.6 & 100.0 & 99.9 \\
        Class6  & 99.6 & 98.5 & 99.7 & 99.2 \\
        Class7  & 94.6 & 93.4 & 100.0 & 100.0 \\
        Class8  & 98.4 & 96.9 & 97.6 & 92.9 \\
        Class9  & 99.6 & 98.3 & 98.9 & 95.9 \\
        Class10 & 99.4 & 98.2 & 99.7 & 98.2 \\
    \midrule
        \textbf{Mean} & \textbf{97.3} & \textbf{94.1} & \textbf{99.0} & \textbf{96.5} \\
    \bottomrule
    \end{tabular}
    }
    \label{tab:dagm_results}
\end{table}

\begin{table}[htbp]
    \centering
    \caption{Performance on the MPDD dataset (class-wise results).}
    \setlength{\tabcolsep}{4pt}
    \renewcommand{\arraystretch}{1.15}
    \resizebox{\columnwidth}{!}{
    \begin{tabular}{lcccc}
    \toprule
        \multirow{2}{*}{\textbf{Objects}} 
        & \multicolumn{2}{c}{\textbf{Pixel}} 
        & \multicolumn{2}{c}{\textbf{Image}} \\
        \cmidrule(lr){2-3} \cmidrule(lr){4-5}
        & \textbf{AUROC} & \textbf{AUPRO} & \textbf{AUROC} & \textbf{AP} \\
    \midrule
        bracket\_black & 97.9 & 93.0 & 79.0 & 88.5 \\
        bracket\_brown & 94.1 & 86.4 & 57.8 & 73.9 \\
        bracket\_white & 99.6 & 97.5 & 91.2 & 90.6 \\
        connector      & 96.9 & 88.6 & 83.6 & 79.1 \\
        metal\_plate   & 94.4 & 87.0 & 66.9 & 86.8 \\
        tubes          & 98.5 & 94.0 & 96.3 & 98.4 \\
    \midrule
        \textbf{Mean}  & \textbf{96.9} & \textbf{91.1} & \textbf{79.1} & \textbf{86.2} \\
    \bottomrule
    \end{tabular}
    }
    \label{tab:mpdd_results}
\end{table}

% \begin{table}[htbp]
%     \centering
%     \caption{Performance on the BTAD dataset (class-wise results).}
%     \setlength{\tabcolsep}{4pt}
%     \renewcommand{\arraystretch}{1.15}
%     \resizebox{\columnwidth}{!}{
%     \begin{tabular}{lcccc}
%     \toprule
%         \multirow{2}{*}{\textbf{Objects}} 
%         & \multicolumn{2}{c}{\textbf{Pixel}} 
%         & \multicolumn{2}{c}{\textbf{Image}} \\
%         \cmidrule(lr){2-3} \cmidrule(lr){4-5}
%         & \textbf{AUROC} & \textbf{AUPRO} & \textbf{AUROC} & \textbf{AP} \\
%     \midrule
%         01   & 94.5 & 78.9 & 95.4 & 98.4 \\
%         02   & 95.4 & 66.3 & 85.5 & 97.7 \\
%         03   & 97.5 & 94.9 & 98.6 & 93.1 \\
%     \midrule
%         \textbf{Mean} & \textbf{95.8} & \textbf{80.0} & \textbf{93.2} & \textbf{96.4} \\
%     \bottomrule
%     \end{tabular}
%     }
%     \label{tab:btad_results}
% \end{table}

\begin{table}[htbp]
    \centering
    \caption{Performance on the BTAD dataset (class-wise results).}
    \setlength{\tabcolsep}{4pt}
    \renewcommand{\arraystretch}{1.15}
    \resizebox{\columnwidth}{!}{
    \begin{tabular}{lcccc}
    \toprule
        \multirow{2}{*}{\textbf{Objects}} 
        & \multicolumn{2}{c}{\textbf{Pixel}} 
        & \multicolumn{2}{c}{\textbf{Image}} \\
        \cmidrule(lr){2-3} \cmidrule(lr){4-5}
        & \textbf{AUROC} & \textbf{AUPRO} & \textbf{AUROC} & \textbf{AP} \\
    \midrule
        01   & 94.5 & 78.9 & 95.4 & 98.4 \\
        02   & 95.4 & 66.3 & 85.5 & 97.7 \\
        03   & 97.5 & 94.9 & 98.6 & 93.1 \\
    \midrule
        \textbf{Mean} & \textbf{95.8} & \textbf{80.0} & \textbf{93.2} & \textbf{96.4} \\
    \bottomrule
    \end{tabular}
    }
    \label{tab:btad_results}
\end{table}

\begin{table}[htbp]
    \centering
    \caption{Performance on the DTD dataset (class-wise results).}
    \setlength{\tabcolsep}{4pt}
    \renewcommand{\arraystretch}{1.15}
    \resizebox{\columnwidth}{!}{
    \begin{tabular}{lcccc}
    \toprule
        \multirow{2}{*}{\textbf{Objects}} 
        & \multicolumn{2}{c}{\textbf{Pixel}} 
        & \multicolumn{2}{c}{\textbf{Image}} \\
        \cmidrule(lr){2-3} \cmidrule(lr){4-5}
        & \textbf{AUROC} & \textbf{AUPRO} & \textbf{AUROC} & \textbf{AP} \\
    \midrule
        Blotchy\_099    & 99.2 & 95.3 & 99.9 & 100.0 \\
        Fibrous\_183    & 99.4 & 97.7 & 100.0 & 100.0 \\
        Marbled\_078    & 99.0 & 96.0 & 99.6 & 99.9 \\
        Matted\_069     & 98.2 & 83.1 & 97.2 & 99.3 \\
        Mesh\_114       & 97.3 & 86.7 & 92.0 & 96.8 \\
        Perforated\_037 & 95.9 & 89.6 & 95.7 & 99.0 \\
        Stratified\_154 & 99.5 & 92.6 & 99.8 & 100.0 \\
        Woven\_001      & 99.8 & 98.9 & 100.0 & 100.0 \\
        Woven\_068      & 98.8 & 91.9 & 94.6 & 96.9 \\
        Woven\_104      & 98.2 & 93.3 & 99.9 & 100.0 \\
        Woven\_125      & 99.5 & 93.4 & 100.0 & 100.0 \\
        Woven\_127      & 95.2 & 93.1 & 96.1 & 97.4 \\
    \midrule
        \textbf{Mean}   & \textbf{98.3} & \textbf{92.6} & \textbf{97.9} & \textbf{99.1} \\
    \bottomrule
    \end{tabular}
    }
    \label{tab:DTD}
\end{table}

\begin{table}[htbp]
    \centering
    \caption{Pixel- and image-level anomaly detection performance on MVTec dataset using FB-CLIP.}
    \setlength{\tabcolsep}{4pt}
    \renewcommand{\arraystretch}{1.15}
    \resizebox{\columnwidth}{!}{
    \begin{tabular}{lcccc}
    \toprule
        \multirow{2}{*}{\textbf{Objects}} 
        & \multicolumn{2}{c}{\textbf{Pixel}} 
        & \multicolumn{2}{c}{\textbf{Image}} \\
        \cmidrule(lr){2-3} \cmidrule(lr){4-5}
        & \textbf{AUROC} & \textbf{AUPRO} & \textbf{AUROC} & \textbf{AP} \\
    \midrule
        bottle       & 93.4  & 86.7  & 96.3  & 98.8 \\
        cable        & 79.4  & 69.2  & 73.9  & 84.1 \\
        capsule      & 96.2  & 91.8  & 93.3  & 98.6 \\
        carpet       & 99.4  & 94.6  & 100.0 & 100.0 \\
        grid         & 97.6  & 80.7  & 99.0  & 99.7 \\
        hazelnut     & 98.1  & 88.7  & 93.2  & 96.6 \\
        leather      & 99.4  & 97.9  & 100.0 & 100.0 \\
        metal\_nut   & 63.3  & 72.4  & 75.1  & 94.7 \\
        pill         & 91.3  & 94.2  & 87.2  & 97.4 \\
        screw        & 98.6  & 92.8  & 86.9  & 95.1 \\
        tile         & 97.9  & 91.6  & 98.8  & 99.5 \\
        toothbrush   & 92.0  & 89.3  & 95.3  & 98.2 \\
        transistor   & 75.6  & 56.2  & 89.0  & 87.0 \\
        wood         & 98.4  & 95.8  & 98.7  & 99.6 \\
        zipper       & 97.3  & 83.5  & 98.6  & 99.6 \\
    \midrule
        \textbf{Mean} & \textbf{91.9} & \textbf{85.7} & \textbf{92.4} & \textbf{96.6} \\
    \bottomrule
    \end{tabular}
    }
    \label{tab:visa_to_mvtec}
\end{table}

\begin{table}[htbp]
    \centering
    \caption{Cross-domain zero-shot performance from MVTec to VISA dataset.}
    \setlength{\tabcolsep}{4pt}
    \renewcommand{\arraystretch}{1.15}
    \resizebox{\columnwidth}{!}{
    \begin{tabular}{lcccc}
    \toprule
        \multirow{2}{*}{\textbf{Objects}} 
        & \multicolumn{2}{c}{\textbf{Pixel}} 
        & \multicolumn{2}{c}{\textbf{Image}} \\
        \cmidrule(lr){2-3} \cmidrule(lr){4-5}
        & \textbf{AUROC} & \textbf{AUPRO} & \textbf{AUROC} & \textbf{AP} \\
    \midrule
        candle      & 98.8 & 96.3 & 92.1 & 93.5 \\
        capsules    & 97.3 & 91.0 & 94.2 & 97.0 \\
        cashew      & 96.4 & 97.1 & 94.3 & 97.5 \\
        chewinggum  & 99.4 & 94.0 & 98.6 & 99.4 \\
        fryum       & 95.9 & 93.6 & 97.0 & 98.7 \\
        macaroni1   & 99.5 & 96.7 & 89.5 & 90.5 \\
        macaroni2   & 98.8 & 90.9 & 76.0 & 76.5 \\
        pcb1        & 93.5 & 86.5 & 83.4 & 83.4 \\
        pcb2        & 92.5 & 80.9 & 81.3 & 80.5 \\
        pcb3        & 89.4 & 82.1 & 71.0 & 74.7 \\
        pcb4        & 95.4 & 91.0 & 96.8 & 96.7 \\
        pipe\_fryum & 98.1 & 96.2 & 99.4 & 99.7 \\
    \midrule
        \textbf{Mean} & \textbf{96.3} & \textbf{91.4} & \textbf{89.5} & \textbf{90.7} \\
    \bottomrule
    \end{tabular}
    }
    \label{tab:mvtec_to_visa}
\end{table}

\begin{table}[htbp]
\centering
\caption{Cross-domain zero-shot performance on Real-IAD dataset.}
\setlength{\tabcolsep}{4pt}
\renewcommand{\arraystretch}{1.15}
\resizebox{\columnwidth}{!}{
\begin{tabular}{lcccc}
\toprule
\multirow{2}{*}{\textbf{Objects}} 
& \multicolumn{2}{c}{\textbf{Pixel}} 
& \multicolumn{2}{c}{\textbf{Image}} \\
\cmidrule(lr){2-3} \cmidrule(lr){4-5}
& \textbf{AUROC} & \textbf{AUPRO} & \textbf{AUROC} & \textbf{AP} \\
\midrule
pcb & 95.8 & 82.0 & 69.2 & 76.9 \\
phone\_battery & 77.7 & 94.9 & 85.1 & 84.1 \\
sim\_card\_set & 99.8 & 98.5 & 95.6 & 96.3 \\
switch & 88.4 & 79.5 & 69.4 & 75.0 \\
terminalblock & 97.9 & 93.1 & 86.2 & 88.8 \\
toothbrush & 93.9 & 82.9 & 71.4 & 76.9 \\
bottle\_cap & 98.3 & 92.8 & 78.5 & 78.8 \\
end\_cap & 93.3 & 76.4 & 71.7 & 77.0 \\
fire\_hood & 99.2 & 95.5 & 83.3 & 75.2 \\
mounts & 97.4 & 93.0 & 83.6 & 68.8 \\
plastic\_nut & 96.7 & 84.0 & 82.7 & 73.4 \\
plastic\_plug & 98.3 & 94.0 & 82.5 & 77.8 \\
regulator & 94.1 & 73.0 & 63.5 & 42.8 \\
rolled\_strip\_base & 99.2 & 97.4 & 93.8 & 96.9 \\
tape & 98.7 & 94.4 & 94.6 & 93.8 \\
porcelain\_doll & 99.4 & 97.1 & 93.5 & 90.2 \\
mint & 94.3 & 84.5 & 77.8 & 78.4 \\
eraser & 99.6 & 96.5 & 89.4 & 89.2 \\
button\_battery & 97.9 & 89.3 & 79.0 & 84.4 \\
toy & 83.5 & 76.2 & 76.8 & 84.0 \\
transistor1 & 94.6 & 78.1 & 74.6 & 80.6 \\
usb & 95.4 & 83.8 & 70.6 & 69.9 \\
usb\_adaptor & 98.8 & 93.5 & 81.2 & 74.9 \\
zipper & 96.5 & 90.8 & 91.0 & 94.8 \\
toy\_brick & 98.2 & 91.1 & 79.7 & 74.8 \\
u\_block & 98.9 & 93.5 & 80.4 & 64.6 \\
vcpill & 97.6 & 84.9 & 85.3 & 83.8 \\
wooden\_beads & 98.3 & 87.5 & 80.3 & 75.1 \\
woodstick & 98.2 & 90.3 & 85.0 & 74.6 \\
audiojack & 95.8 & 77.8 & 63.2 & 48.8 \\
\midrule
\textbf{Mean} & \textbf{95.9} & \textbf{88.2} & \textbf{80.6} & \textbf{78.4} \\
\bottomrule
\end{tabular}}
\label{tab:realIAD_results}
\end{table}

%% file: sec/LimitationLimitation.tex
In this section, we analyze FB-CLIP from two perspectives: the inference stage and failure detection cases.

\begin{table}[ht]
\centering

\caption{Comparison with recent SOTA methods in the inference process.}
\label{tab:infer}
\resizebox{\linewidth}{!}{  % Resize the table to fit the column width
\begin{tabular}{ccccc}
\toprule
 \textcolor{orange}{Infer Metric}
 & AF-CLIP & FAPrompt & FB-CLIP \\
\midrule
Time & 112 ms &  232 ms
 & 215 ms \\
Memory & 2.5G &  2.6G
 & 3.3G \\
\bottomrule
\end{tabular}
}
\end{table}

\begin{figure}[htbp]
    \centering
    \includegraphics[width=0.5\textwidth]{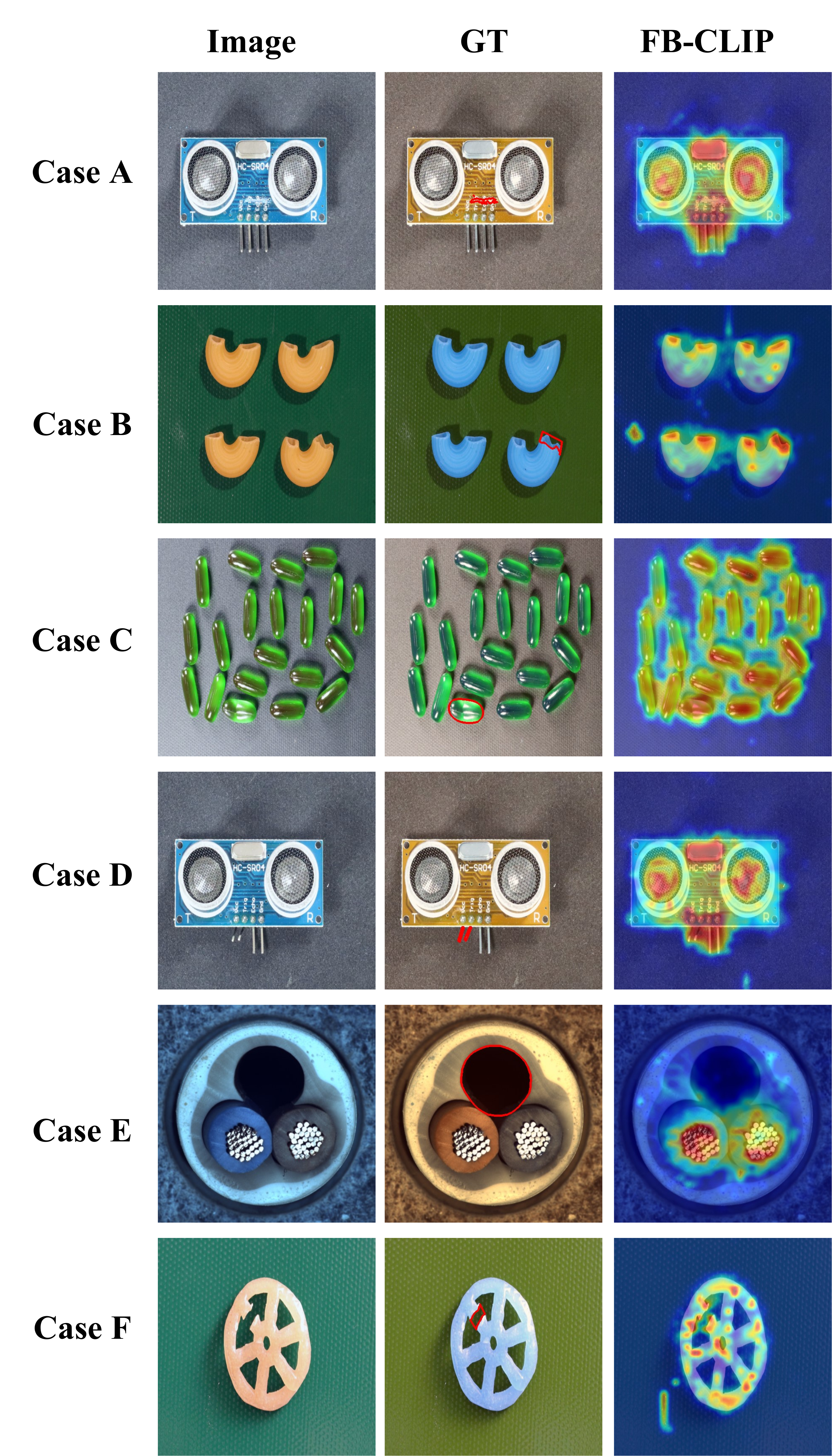} % 修改为你的图片路径
    \caption{
 Visualization results of inaccurate anomaly localization using the FB-CLIP method.
    }
    \label{fig:fbclip_bad_case}
\end{figure}

In Figure ~\ref{fig:fbclip_bad_case}, we present several representative bad cases of our FB-CLIP model to illustrate its limitations. 
Note that these examples are selected failure samples, and similar issues may also exist in other anomaly detection methods.

For the sensor module in Cases A and D, FB-CLIP over-responds to non-functional variations such as material and color changes on the PCB, while failing to capture subtle structural defects on the pins. 
In Case B (the crescent-shaped object), the model produces diffused activation maps and cannot localize fine-grained edge damages. 
In Case C (dense capsules), the method fails to distinguish the abnormal individual from the cluster, as anomalies are overwhelmed by surrounding similar instances. 
For Case E (the pipeline hole), FB-CLIP incorrectly treats texture and color variations inside the hole as anomalies, with poorly localized boundaries. 
In Case F (the wheel component), the activation spreads over the entire object rather than concentrating on the defective spoke region, leading to inaccurate defect localization.

These selected cases demonstrate that FB-CLIP still faces challenges in distinguishing functional defects from appearance variations, detecting tiny local anomalies, and maintaining precise localization under complex backgrounds. 
However, such limitations are commonly observed in many feature-based anomaly detection approaches rather than being unique to our method.

Beyond the above qualitative limitations, FB-CLIP also has shortcomings in computational efficiency, as shown in Table \ref{tab:infer}.
Although achieving competitive detection performance, FB-CLIP consumes more memory (3.3G) and longer inference time (215 ms) than AF-CLIP, which restricts its deployment in resource-constrained real-time scenarios.
These observations indicate that the performance improvement of FB-CLIP is accompanied by higher computational costs, and there is still room for optimization in efficiency.

%% file: sec/Occusion.tex
\begin{figure}[htbp]
    \centering
    \includegraphics[width=0.5\textwidth]{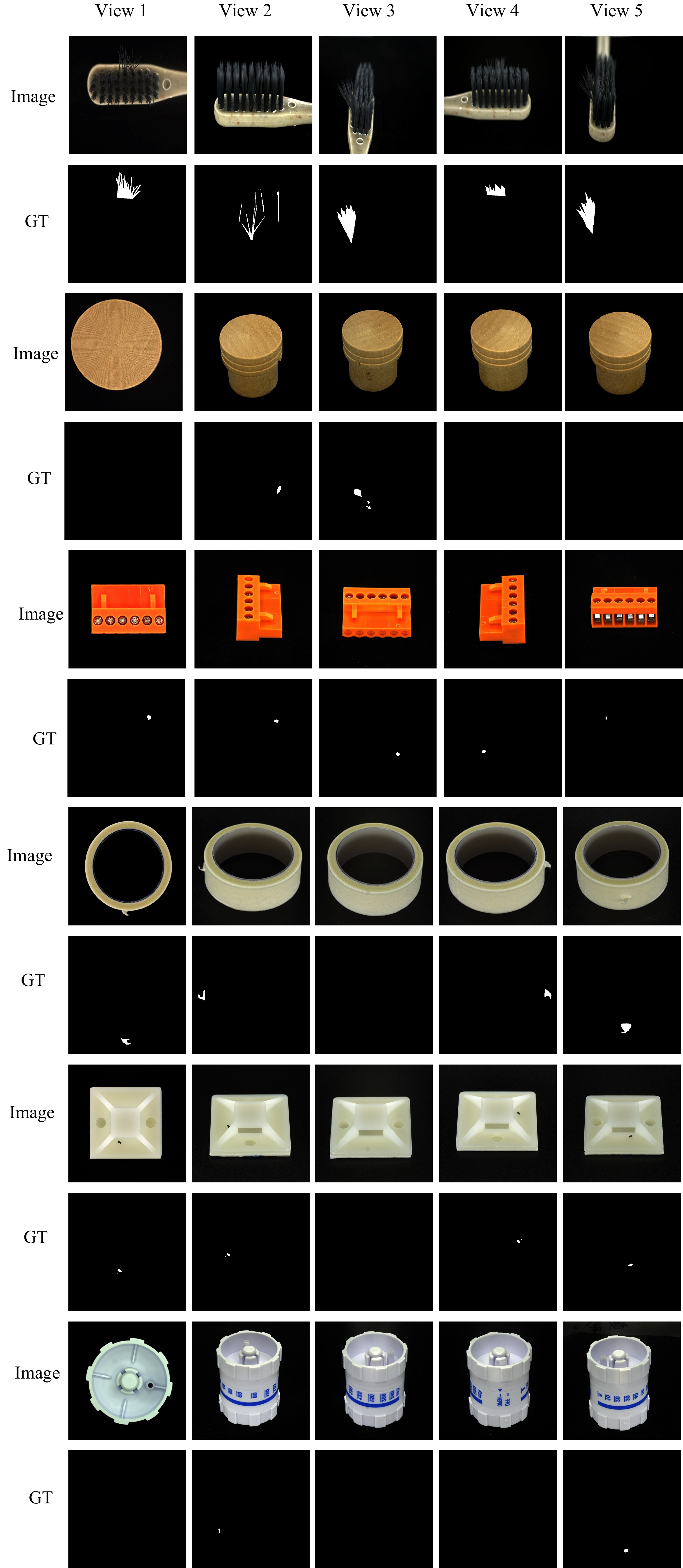} % 修改为你的图片路径
    \caption{
    Visualization of the same object from different viewpoints in the Real-IAD dataset.
    }
    \label{fig:occu}
\end{figure}

While existing methods demonstrate high performance on standard benchmarks, anomaly detection in more challenging real-world scenarios—such as those involving physical occlusion or partial blockage—remains significantly difficult. As shown in Figure~\ref{fig:occu}, physical occlusion can obscure critical regions, making it hard for models to capture complete feature representations, thereby reducing detection and localization accuracy.

In such complex scenarios, models are required not only to distinguish subtle differences between normal and abnormal objects but also to infer information about occluded regions. This demands approaches that can effectively integrate local and global features to enhance perception of partially visible targets. Techniques such as multi-view information fusion or context-based feature completion may offer effective solutions for these challenges.

Extending anomaly detection methods to explicitly handle physically occluded scenarios not only tests the robustness of existing models but also motivates the design of more resilient multi-strategy fusion and multi-view perception frameworks.

\textbf{Inspired by the Real-IAD Dataset and the suggestions of the reviewers of this paper, we argue that another important application scenario for anomaly detection is leveraging multi-view observations to identify and localize anomalies. In practical environments, objects may be partially occluded, or critical regions may not be visible from a single viewpoint. As a result, relying solely on single-view information can limit the model’s ability to accurately detect and localize abnormal patterns. By integrating complementary information from multiple viewpoints, models can obtain richer and more complete feature representations, which helps alleviate the impact of occlusion and incomplete observations. Therefore, designing anomaly detection approaches that effectively utilize multi-view information is a promising direction for improving robustness and reliability in complex real-world scenarios.}

